\documentclass[a4paper,fleqn]{cas-dc}

\usepackage[authoryear,longnamesfirst]{natbib}

\def\tsc#1{\csdef{#1}{\textsc{\lowercase{#1}}\xspace}}
\tsc{WGM}
\tsc{QE}

\usepackage[figuresright]{rotating}
\usepackage{cleveref}
\usepackage{graphicx}
\usepackage{tikz}
\usepackage{color}
\usepackage{enumitem}
\usepackage{tabularx} 
\usepackage{tikz}
\usepackage{stfloats}
\usetikzlibrary{bayesnet}

\tikzset{>=latex}

\usepackage{pgf}
\usepackage{threeparttable}
\usepackage{multirow}
\usepackage{listings}
\usepackage{amsfonts}
\usepackage{amsmath,amssymb}
\usepackage{makecell}
\usepackage{adjustbox}
\usepackage{graphics}
\usepackage{algorithm}
\usepackage{algorithmic} %format of the algorithm
\usepackage{subfigure}
\usepackage{tablefootnote} % for table footnotes
\usepackage{multibib}
\newcites{AP}{References}

\usepackage{subfigure}
\usepackage{import}
\usepackage{booktabs}
\usepackage{fix-cm}

\usepackage[accsupp]{axessibility}  % Improves PDF readability for those with disabilities.

\begin{document}

\let\WriteBookmarks\relax
\def\floatpagepagefraction{1}
\def\textpagefraction{.001}

\def\tablefootnotemark#1{\textsuperscript{\hyperref[#1]{\getrefnumber{#1}}}}

\shorttitle{AGAD: Adversarial Generative Anomaly Detection}    

\shortauthors{J. Shi and N. Zhang}  

\title [mode = title]{AGAD: Adversarial Generative Anomaly Detection}  

% \tnotemark[<tnote number>] 



\let\printorcid\relax % Remove ORCID footnote

\author[]{Jian Shi}
\ead{shi_jian@nec.cn}
\author[]{Ni Zhang}
\ead{zhangni_nlc@nec.cn}

% \cormark[2]


% \ead[url]{<URL>}

% \credit{<Credit authorship details>}

\affiliation[]{organization={NEC Laboratories China},
            city={Beijing},
%          citysep={}, % Uncomment if no comma needed between city and postcode
            country={China}}

% Footnote of the second author



% \affiliation[<aff no>]{organization={},
% %          citysep={}, % Uncomment if no comma needed between city and postcode
%             country={China}}

% \cortext[1]{Corresponding author}



\begin{abstract}
Anomaly detection suffered from the lack of anomalies due to the diversity of abnormalities and the difficulties of obtaining large-scale anomaly data. Semi-supervised anomaly detection methods are often used to solely leverage normal data to detect abnormalities that deviated from the learnt normality distributions. Meanwhile, given the fact that limited anomaly data can be obtained with a minor cost in practice, some researches also investigated anomaly detection methods under supervised scenarios with limited anomaly data. In order to address the lack of abnormal data for robust anomaly detection, we propose Adversarial Generative Anomaly Detection (AGAD), a self-contrast-based anomaly detection paradigm that learns to detect anomalies by generating \textit{contextual adversarial information} from the massive normal examples. Essentially, our method generates pseudo-anomaly data for both supervised and semi-supervised anomaly detection scenarios. Extensive experiments are carried out on multiple benchmark datasets and real-world datasets, the results show significant improvement in both supervised and semi-supervised scenarios. Importantly, our approach is data-efficient that can boost up the detection accuracy with no more than 5\% anomalous training data.
\end{abstract}


% \begin{highlights}
% \item We propose a simple but effective anomaly detection paradigm that unifies both supervised and semi-supervised anomaly detection schemes, and can leverage the existence of small anomaly data for much improved performance.

% \item We introduce \textit{contextual adversarial information}, for better learning discriminative features between normal and abnormal data by generating pseudo-anomaly data in a contrastive manner, obtaining results that significantly improved over the current state-of-the-art by a large margin.
% % \item {\bfseries Versatility}: 
% % \item {\bfseries Data-efficient supervised anomaly detection}: 
% \item Our method is data-efficient, that can effectively be boosted up model performances with no more than 5\% anomaly images.

% \item Apart from common benchmarking datasets, we further evaluated our method on various real-world medical datasets, that can achieve high detection accuracy.
% \end{highlights}

% Keywords

\maketitle


\section{Introduction}

Anomaly detection aims at detecting exceptional data instances that significantly deviated from the normality data, which has an increasing demands in medical diagnosis, fraud detection and many other fields. In most tasks, anomalous data is scarce and diverse so that anomaly detection is commonly modeled as semi-supervised\footnote{\scriptsize Some early studies refer the training with only normal data as unsupervised anomaly detection, but unsupervised approaches shall be formed as class-agnostic without the prior knowledge of the dataset labels. To avoid unnecessary confusion, we refer such strategy as semi-supervised anomaly detection, following \cite{Musa2021,Pang2021-ch}.} learning problems, in which anomalous data is mostly considered as not available during the training and the training data contains only "normal" class. Traditional anomaly detection methods mainly include one-class methods (e.g. One-class SVM), reconstruction-based methods (e.g. AutoEncoders) and statistical models (e.g. KMeans). Recently, this field attracted more attention due to the marriage with Generative Adversarial Networks (GANs) \cite{NIPS2014_GAN}, including \cite{sabokrou2018adversarially,pmlr-v80-ruff18a,anogan-1703.05921,Akcay2019-jx,Akcay2019-tm,Zhao2018,DeepDisa-2202.00050}. However, most anomaly detection methods suffer from low-recall rate that many normal samples are wrongly reported as anomalies while true yet sophisticated anomalies are missed \cite{Pang2021-ch}. In this field, reconstruction-based anomaly detection is a classic semi-supervised anomaly detection method, that hypothesises the anomaly data is harder to be reconstructed if only learns to reconstruct normal data. Though many studies \cite{Akcay2019-tm,GeoTrans1805.10917,Ye2022} presented promising results with reconstruction-based anomaly detection models, it cannot guarantee those models are optimized towards better anomaly detection since they are trained for reconstruction. Zooming out, common semi-supervised anomaly detection schemes neglected the fact that limited anomaly data is obtainable in most anomaly detection tasks without much efforts. Thus, supervised anomaly detection is also researched. Specifically, fully supervised anomaly detection is normally referred as imbalanced classification tasks \cite{Kong2020improving,Kim2020-rq} that requires the both availabilities of labelled normal and abnormal data, while a few works \cite{Daniel2019DeepVS,wu2021surrogate} focus on weakly-supervised anomaly detection that allows learning within a partially labelled dataset. Nevertheless, due to the nature of anomalies, the collected anomaly data can hardly cover all anomaly types \cite{deepweak1910.13601}.  Data efficient methods are therefore beneficial for anomaly detection in practice use.
% In this sense, this work aims at improving the data-efficiency for anomaly detection tasks.

% attempted using GAN-based anomaly detection methods to examine discriminative features between normality and pseudo-anomaly data.

% well-reconstruct the data from the training distribution but not for the unseen abnormalities.

% In general, reconstruction-based methods hypothesize strong distribution differences between normality and abnormality data, and it is used as a typical surrogate task for anomaly detection.

% In this sense, a preferred approach is to ensure the model reconstructs normal images better while failing the reconstruction of abnormal images.

% Apparently, collecting large-scale abnormalities in real-world is impractical due to the scarcity of anomalies.
Inspired by the recent success of contrastive learning \cite{chen2020simple,chen2020big,he2019moco,chen2020mocov2,grill2020bootstrap,chen2020exploring,caron2020unsupervised}, we tend to integrate generative and contrastive methods to generate pseudo-anomaly data to address the lack of abnormalities. In such sense, we propose AGAD, Adversarial Generative Anomaly Detection, a novel reconstruction-based anomaly detection method that generates \textit{contextual adversarial information} from normal data contrastively as well as learning from abnormalities. To be specific, we introduced \textit{contextual adversarial information}, which is essentially the discriminative features of pseudo-anomaly data against normalities. We would hereby expect the model can further enlarge the distances between normal and abnormal distributions. Key contributions of this work include:
\begin{itemize}[leftmargin=*]
    % \item {\bfseries Anomaly-aware anomaly detection}: 
    \item introduce \textit{contextual adversarial information}, for better learning discriminative features between normal and abnormal data by generating pseudo-anomaly data in a contrastive manner, obtaining results that significantly improved over the current state-of-the-art by a large margin.
    % \item {\bfseries Versatility}: 
    \item propose a simple yet effective anomaly detection paradigm that unifies both supervised and semi-supervised anomaly detection schemes, and can leverage the existence of small anomaly data for much improved performance.
    % \item {\bfseries Data-efficient supervised anomaly detection}: 
    \item data-efficient, the proposed method can effectively be boosted up model performances with no more than 5\% anomaly images.
\end{itemize}

\section{Related Works}

As surveys \cite{Pang2021-ch,BeulaRani2020} indicated, semi-supervised anomaly detection methods dominated in this research field. Recently, with the introduction of GANs \cite{NIPS2014_GAN}, many researches attempted to bring GANs into anomaly detection. Earlier works like AnoGAN \cite{Schlegl2017-zr} learns the normal data distributions with GANs and attempts to reconstruct the most similar images by optimizing a latent noise vector iteratively. With the success of Adversarial Auto Encoders (AAE) \cite{AAE2016}, some more recent works combined AutoEncoders and GANs together to detect anomalies. Particularly, GANomaly \cite{Akcay2019-jx} further regularized the latent spaces between inputs and reconstructed images, then some following works improved it with more advanced generators like UNet \cite{Akcay2019-tm} and UNet++ \cite{Cheng2020}. Most of those works attempted to achieve better anomaly detection performances by learning towards the better reconstruction of normalities, disregarding the awareness of abnormalities.

% EGBAD \cite{EGBAD1802.06222} immerged a BiGAN \cite{BiGAN-Ding2020-cr} -like structure into anomaly detection tasks, then \cite{BiGANADKaplan2020-pd} improved the training logic towards a better performance.
In contrast, supervised anomaly detection methods take both normal and abnormal data into account. Notably, supervised anomaly detection was mostly formulated as an imbalanced classification problem that addressed with different classification approaches \cite{Agrawal2015} or sampling strategies \cite{mishra2017handling,Gonzalez}. However, supervised anomaly detection is normally in course of little labelled or noisy labelled data with limited supervisions. In such sense, Deep SAD \cite{ruff2020deep}, a general method based on Deep SVDD \cite{pmlr-v80-ruff18a}, proposed a two-stage training with information-theoretic framework. TLSAD \cite{Feng2021} further consolidated the model’s discriminative power with a transfer learning framework, which relied on an additional large-scale reference dataset for the model training. Recent advances \cite{Razakarivony2014,Lbbering2020,Yamanaka2019} took the advantage of reconstruction-based anomaly detection framework and proposed different methods to address the unbalanced classification problem by minimizing the reconstruction error for normal data, and to maximizing it for anomalies. ESAD \cite{ESAD2020} considers both ideas of information-theoretic framework and reconstruction-based anomaly detection framework to optimize mutual information and entropy. Due to the limited availability of anomaly samples, a major challenge of supervised anomaly detection is to improve the data-efficiency of learning algorithms.

With the successes of supervised deep learning models, more researches \cite{Dosovitskiy2014Discriminative,wen2016discriminative} explored the methods of reducing the needs of labelled samples, as known as discriminative feature learning. Learning discriminative features has also proven to be effective in anomaly detection field, GeoTrans \cite{GeoTrans1805.10917} leveraged geometric transformations to learn discriminative features. ARNet \cite{Ye2022} attempted to use embedding-guided feature restoration to learn more semantic anomaly features. Recently, self-supervised representation learning is getting more attention since it learns domain-specific discriminative features for downstream tasks without any labelling efforts.  Specifically, contrastive learning methods \cite{chen2020simple,chen2020big,he2019moco,chen2020mocov2,grill2020bootstrap,chen2020exploring,caron2020unsupervised} has proven to be one of the most promising approaches in unsupervised representation learning. With the recent works of the integration \cite{tack2020csi,Cho2021} of contrastive learning and anomaly detection tasks, we tended to further combine it with GANs to explore discriminative anomaly features unsupervisedly.

% Autoencoders learns images features by a surrogate reconstruction task, hereby recognized as a classic representation learner. Similar successors \cite{infoGAN2016} utilized GANs to learn disentangled representations with generative methods.

% Alternative to generative models,

% Meanwhile, some researches \cite{supCL2004.11362} adapted contrastive learning for supervised learning as well.

\section{Proposed Method: AGAD}

This section will elaborate the main methodology of our proposed AGAD method. We firstly presented an overview of the problem and our solution, then provided a detailed introduction of the training objectives, along with the training algorithm. Lastly, we compared our method with other relative algorithms.

\subsection{Method Overview} \label{pro_def}

Given an anomaly detection dataset $\mathcal{X}$ with normality $\mathcal{X}_n$ and abnormality $\mathcal{X}_a$, whereas $\mathcal{X}_n=\{x_n: x_n \sim p_n(x)\}$ and $\mathcal{X}_a=\{x_a: x_a \sim p_a(x)\}$.  To distinguish $\mathcal{X}_n$ and $\mathcal{X}_a$, 
% we utilized a generator $G$ with an encoder $G_E$ to encode image data $\mathcal{X}$ into latent space $\mathcal{Z}$, and a decoder $G_D$ to decode latent features $\mathcal{Z}$ into reconstructed data space $\hat{\mathcal{X}}$. 
we utilized a GAN-like structure with a generator $G$ to reconstruct image data $\mathcal{X}$ into reconstructed data space $\hat{\mathcal{X}}$, and a discriminator $D$ is used to distinguish between data domains of $\mathcal{X}$ and $\hat{\mathcal{X}}$. Generally, $G$ tends to generate realistic $\hat{\mathcal{X}}$ that is indistinguishable to $\mathcal{X}$ to fool $D$. Notably, $D$ also acts as a feature extractor to compress image data $\mathcal{X}$ into latent spaces $\mathcal{Z}$, denoted as $D_\mathcal{Z}$. 

Whilst training with normality data $\mathcal{X}_n$, the aforementioned routine process learns to reconstruct $\mathcal{X}_n$ to $\hat{\mathcal{X}_n}$ with minimum differences. Next, anomaly-awareness is achieved by the \textit{contextual adversarial information}. Specifically, we assume the reconstructed $\hat{\mathcal{X}}$ as pseudo-anomaly and expect the reconstruction of $\hat{\mathcal{X}}$ will be failed. Mathematically, given a distance function $\mathcal{F}_{dist}$, we expect to learn a suite of parameter $\theta_{G}$ and $\theta_{D}$ that can achieve:
\begin{equation}
    \min{\mathcal{F}_{dist} (\mathcal{X}_n,\hat{\mathcal{X}_n};\theta_{G})}
    \wedge
    \max{\mathcal{F}_{dist} (\hat{\mathcal{X}_n}, G(\hat{\mathcal{X}_n});\theta_{G})},
\end{equation}
\vspace{-2em}
\begin{equation}
    \min_{G}\max_{D}[\log{D(\mathcal{X}_n;\theta_{D})} + \log{(1-D(G(\mathcal{X}_n};\theta_{G});\theta_{D})].
\end{equation}

\begin{figure*}
    \centering
    \begin{adjustbox}{width=.9\hsize,center}
    % \import{.}{arch.tex}
    \includegraphics{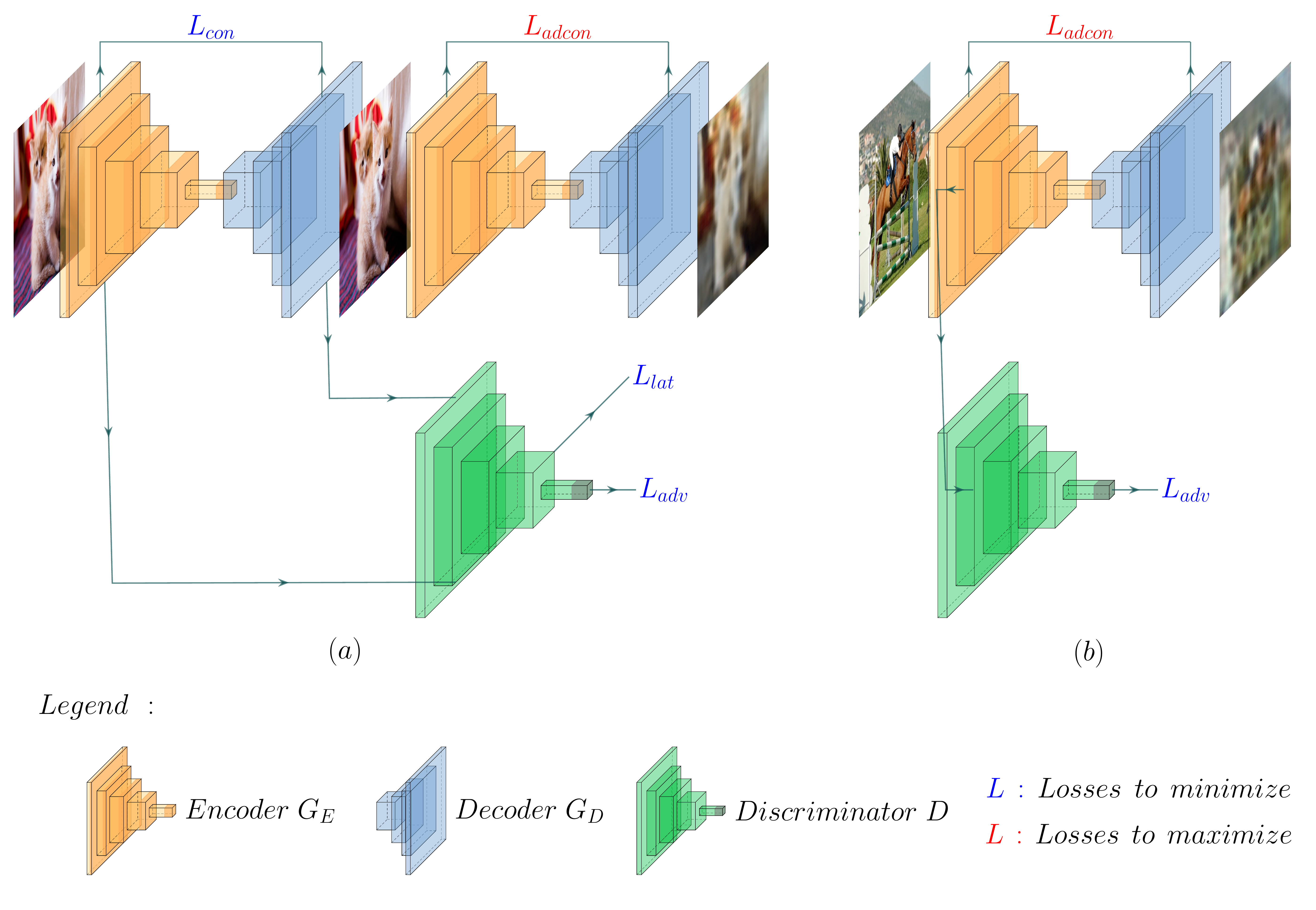}
    \end{adjustbox}
    \caption{The proposed AGAD architecture. (a) training pipeline for normal data: that brings anomaly-awareness by minimizing the contextual loss $L_{con}$ and maximize the contextual adversarial loss $L_{adcon}$. (b) training pipeline for abnormal data: learns to fail the reconstruction of anomaly directly.}
    \label{fig:arch}
\protect\end{figure*}

When abnormal data $\mathcal{X}_a$ available, the generator $G$ learns to fail the reconstruction directly and the discriminator $D$ can distinguish it from the normal:
\begin{equation}
    \max_{\theta_{G}}{\mathcal{F}_{dist} (\mathcal{X}_a, G(\mathcal{X}_a))},
\end{equation}
\vspace{-2em}
\begin{equation}
    \max_{D}[\log{(1-D(\mathcal{X}_a};\theta_{D})].
\end{equation}
The corresponding structure is shown in \Cref{fig:arch}.

\protect

\subsubsection{Contextual Adversarial Information} In general, our consideration is not to expect a strong generator that reconstructs normal data samples perfectly, but enlarges upon the differences between the normality and abnormality data. Essentially, the introduced \textit{contextual adversarial information} leverages pseudo-anomaly features that we generated against normality data, hereby the same generator reconstructs normalities while failed the reconstruction of pseudo-anomaly examples, and therefore, aware the anomaly features.

\subsection{Objective Function Decomposition}

Our method fits both supervised and semi-supervised scenarios and hereby includes a \textit{Normality Loss} for normalities and a \textit{Anomaly Loss} for abnormalities. The detailed composition will be discussed as following.

\noindent\textbf{Adversarial Loss.} We utilized the common adversarial loss \Cref{eq:l_adv} proposed in \cite{NIPS2014_GAN} to ensure that the network $G$ generates realistic reconstructions under the adversarial training with $D$. It is formulated as:
\begin{equation}
    \label{eq:l_adv}
    \mathcal{L}_{adv} = \mathbb{E}_{x\sim p_x}[\log{D(x)}]+ \mathbb{E}_{x\sim p_x}[\log[1-D(\hat{x})]].
\end{equation}

\noindent\textbf{Contextual Loss.} As illustrated by \cite{Akcay2019-tm}, the adversarial loss will not guarantee to learn the contextual information with respect to the input. Specifically, the adversarial loss may produce realistic reconstructions to fool the $D$ sufficiently, but may lose the contextual information that contains original image details. Thus, a $L_1$ loss is used towards a better reconstruction. It is formulated as:
\begin{equation}
    \label{eq:l_con}
    \mathcal{L}_{con} = \mathbb{E}_{x\sim p_x}|x-\hat{x}|_1.
\end{equation}

\noindent\textbf{Latent Loss.} \Cref{eq:l_adv} and \Cref{eq:l_con} minimize $\mathcal{F}_{dist} (\mathcal{X},\hat{\mathcal{X}})$ for producing realistic reconstructions with respect to the input in the pixel-level. Meanwhile, a latent-reconstruction loss \Cref{eq:l_lat} is also used to ensure the solid reconstruction of latent representations. It is formulated as:
\begin{equation}
    \label{eq:l_lat}
    \mathcal{L}_{lat} = \mathbb{E}_{x\sim p_x}|D_\mathcal{Z}(x)-D_\mathcal{Z}(\hat{x})|_2.
\end{equation}

\noindent\textbf{Contextual Adversarial Loss\footnote{\scriptsize though we used a naive negative L1 loss here, the proposed \textit{contextual adversarial loss} can be any function as $W(D(X))$ where $D$ is a distance function and $W$ wraps the distance error, as long as $W$ is monotonically decreasing whilst $D$ is monotonically increasing.}.} In order to differentiate from the normalities, we intend to maximize the reconstruction error for anomaly data, including the true anomalies and the reconstructed pseudo-anomaly data. We formulated \textit{contextual adversarial loss} based on an intuitive theorem that maximizing a function is equivalent to minimize its negative. It is formulated as:
\begin{equation}
    \mathcal{L}_{adcon} = -\mathbb{E}_{x\sim p_x}|\hat{x}-G(\hat{x})|_1.
\end{equation}

\noindent\textbf{Normality Loss.} Normality loss considers improving the reconstruction quality for normal data, as well as failing the reconstruction of the generated counterparts. The final formula is a weighted sum of the losses above, where $\lambda_{adv}$, $\lambda_{con}$, $\lambda_{adcon}$, $\lambda_{lat}$ are the weighting parameters. It is formulated as:
\vspace{-1em}
\begin{equation}
    \mathcal{L}_{n} = \lambda_{adv}\mathcal{L}_{adv} + \lambda_{con}\mathcal{L}_{con} + \lambda_{adcon}\mathcal{L}_{adcon} + \lambda_{lat}\mathcal{L}_{lat}.
\end{equation}

\noindent\textbf{Anomaly Loss.} Opposite to the normality loss, anomaly loss solely considers deteriorating the reconstruction of anomaly data with greater errors, Compared to minimizing the negative loss, we found using reciprocal can lead to a smooth training curve. It is formulated as:
\begin{equation}
    \mathcal{L}_a = \frac{\lambda_{adv}}{\mathcal{L}_{adv}} +
        \frac{\lambda_{adcon}}{-\mathcal{L}_{adcon}} + \frac{\lambda_{lat}}{\mathcal{L}_{lat}}.
\end{equation}

\noindent\textbf{Final Loss.} Our final loss function is formulated as:
\begin{equation}
\begin{aligned}
    \mathcal{L}= y\mathcal{L}_a + (1-y)\mathcal{L}_n, y\in{0,1},
    % \begin{cases}
    %     \mathcal{L}_n, & \text{if } y=0 \\
    %     \mathcal{L}_a, & \text{if } y=1 \\
    % \end{cases}
\end{aligned}
\end{equation}
where $y$ is the groudtruth that indicates if the input data is abnormal or not. We followed 0-normal, 1-abnormal convention here.

\subsection{AGAD Model Pipeline}

Next, we dive into the details of the training procedures for our proposed AGAD model. We firstly introduced the naive network architectures we adopted as our backbone models. Then further discussed the effect of the involvement of anomaly data and presented our training algorithm.
% naive UNet \cite{UNet10.1007/978-3-319-24574-4_28}
\subsubsection{Network Details} The proposed \textit{AGAD} model is designed with a generator and a discriminator. The generator takes images as input and outputs generated images, while the discriminator takes images as inputs and outputs corresponding latent embeddings as well as discriminate results (e.g. real/fake). We adopted a UNet++ \cite{Zhou2020} as generator and a naive VGG-like \cite{Simonyan2015VeryDC} classifier as discriminator. The discriminator consists two ${Conv-BN-LeakyReLU}$\footnote{\scriptsize ${Conv-BN-ReLU}$ represents a sequence of a convolution layer, a batch normalization layer, and a leaky rectified linear unit Layer} blocks and a fully convolution classifier.

\subsubsection{Learning with Limited Anomaly Data}\label{sec:opt}
As indicated by many studies \cite{Bottou2010,Ge2015EscapingFS}, mini-batch is a compromise that injects enough noise to each gradient update, while achieving a relative speedy convergence with less computational resources. Thus, whilst sampling mini-batches for training, we would guarantee there are at least 32 anomaly data sampled to avoid noisy gradient estimations by a small batched training when there is limited anomaly data sampled within the mini-batch. Meanwhile, batch normalization \cite{BN2015} is an essential component in neural networks that normalizes the input features by the learnt feature mean/variance within mini-batches. A previous research \cite{advpropXie2020-md} stated the joint learning on the shared batch normalization layers for original and adversarial examples would lead to a worse performance since the adversarial distribution differs from the original. However, in this context, we speculate that the distribution differences are the critical discriminative features for network to learn. The corresponding ablation study refers to \Cref{sec:bn}.

% separating original and adversarial examples into different distributions can lead to better performance. 

\begin{algorithm}[h]
\caption{Training of AGAD}\label{alg:training}
\begin{algorithmic}[1]
\REQUIRE $S$: a set of images with normal $S_n$ and abnormal $S_a$. $f_\theta$: a model parametrized by $\theta$. $\delta$: threshold to reset $\theta_d$. $\eta$: learning rate.

\ENSURE Anomaly detection model $f_\theta$
% repeat-until loop
\REPEAT

\STATE Read mini-batch $B=\{x_1,x_2,...,x_m\}$
\STATE $\hat{x} = G(x)$, $y, z=D(x)$ \COMMENT{Reconstruct input}
\STATE $\hat{x}^\prime = G(\hat{x})$, $\hat{y}, \hat{z}=D(\hat{x})$ \COMMENT{Reconstruct reconstructed input}
\IF{$x \in S_n$}
    \STATE $L_g=\lambda_{adv}l_{adv}(\hat{y}, 1) + \lambda_{con}l_{con}(x,\hat{x}) + \lambda_{adcon}l_{adcon}(\hat{x},\hat{x}^\prime) + \lambda_{lat}l_{lat}(z,
    \hat{z})$
    \STATE $L_d = l_{adv}(\hat{y}, 0) + l_{adv}(y, 1)$
\ELSIF{$x \in S_a$}
    \STATE $L_g=\lambda_{adv}/l_{adv}(\hat{y}, 0) - \lambda_{adcon}/l_{adcon}(\hat{x},\hat{x}^\prime) + \lambda_{lat}/l_{lat}(z, \hat{z})$
    \STATE $L_d = l_{adv}(y, 0)$
\ENDIF
\STATE $\theta_g \Leftarrow \theta_g - \eta \Delta_{\theta_d}L_g$ \COMMENT{Update NetG by stochastic gradient}
\STATE $\theta_d \Leftarrow \theta_d - \eta \Delta_{\theta_d}L_d$ \COMMENT{Update NetD by stochastic gradient}
% \IF{$L_d < \delta$}
%     \STATE Re-initialize $\theta_d$
% \ENDIF
\UNTIL{training finished}

\end{algorithmic}
\end{algorithm}

\subsubsection{Anomaly Score} During the inference time, we used an intuitive scoring function based on the reconstruction quality under image domain to pick out anomaly data. The scoring function is defined as $A(x) = ||x-\hat{x}||_2$, where $|| \cdot ||_2$ represents the $L_2$ norm, $x$ and $\hat{x}$ represent the input data and the reconstructed data $G(x)$ respectively.

\subsection{Relation to Other Algorithms}

\textbf{GANomaly \cite{Akcay2019-jx} and Skip-GANomaly \cite{Akcay2019-tm}} are semi-supervised anomaly detection methods that not only reconstruct input images, but also try to regularize the latent features between the input images and reconstructed images. Specifically, Skip-GANomaly extended GANomaly by replacing the generator with a UNet model, resulting in performance gain. Our proposed method extends the GANomaly framework and makes use of the generated pseudo-anomaly as \textit{contextual adversarial information} to improve anomaly detection performance. 

\textbf{GAN-based supervised anomaly detection \cite{Kim2020-rq,mishra2017handling,Gonzalez}} considers supervised anomaly detection in an imbalanced dataset scenario using reconstruction-based anomaly detection methods. Our method had similar ideas to improve the reconstruction performance for normalities while impair the reconstruction performance for abnormalities. Our strength is to fit both supervised and semi-supervised scenarios, and can leverage generated pseudo-anomaly together with the existence of true anomaly data to improve anomaly detection performance.

\section{Benchmarks}\label{sec:exp}

In this section, we present major experiment results to demonstrate the effectiveness of our proposed AGAD model. We elaborate the public datasets we adopted, then compare our models with several semi-supervised and supervised anomaly detection benchmark methods respectively. Finally, we further evaluate the qualitative results of our method.

\subsection{Benchmark Datasets}
\label{sec:data}
We experiment on benchmark datasets including MNIST, Fashion-MNIST, CIFAR-10 and CIFAR-100. A short briefing of each dataset and preprocessing methods are presented as following.% Moreover, experiments for real-world medical datasets are in Appendix \Cref{app:realworld}.

\begin{itemize}[noitemsep,leftmargin=*]
    \item{\bfseries MNIST} is a handwritten digits with 10 equally distributed classes that has a training set of 60,000 examples, and a test set of 10,000 examples. We resized the images to 32x32 in all experiments.
    
    \item{\bfseries Fashion-MNIST} is a dataset of Zalando’s article images -- consisting of a training set of 60,000 examples and a test set of 10,000 examples. Each example is a 28x28 grayscale image, associated with a label from 10 classes. We resized the images to 32x32 in all experiments.
    
    \item{\bfseries CIFAR-10} consists of 60,000 32x32 color images in 10 equally distributed classes with 6,000 images per class, including 5,000 training images and 1,000 test images.
    
    \item{\bfseries CIFAR-100} similar to CIFAR-10, except with 100 classes containing 600 images each. There are 500 training images and 100 testing images per class. The 100 classes in the CIFAR-100 are grouped into 20 superclasses. Each image comes with a "fine" label (the class to which it belongs) and a "coarse" label (the superclass to which it belongs), which we use in the experiments.
\end{itemize}

% \begin{figure}[h]
%     \centering
%     \caption{Dataset description}\label{tab1}
%     \begin{tabular}{|l|l|l|l|l|}
%     \hline
%     Imaging Protocol & Dataset &  Training (Normal) & Testing (Normal/Case)\\
%     \hline
%     CT              & ChestXray \cite{Kermany2018-eu} &  1341 & 242 / 398 \\
%     Microscopy      & Colon Histopathology  \cite{Lungcolondataset}  &  4500 & 500 / 500 \\
%     Microscopy      & Lung Histopathology  \cite{Lungcolondataset}   &  4500 & 500 / 1000 \\
%     OCT             & Retinal OCT     \cite{Kermany2018-eu}        &  26315 & 250 / 750 \\
%     MRI             & Alzheimer's Dataset     & 2560  & 640 / 639 \\
%     \hline
%     \end{tabular}
% \end{figure}

% \subsection{Metrics}

% \begin{itemize}
%     \item AUROC
% \end{itemize}

\subsection{Results}\label{sec:benchmark}

Several benchmarks are performed in this section with a common metric of AUROC (Area Under the Curve of Receiver Operating Characteristics), shown in \% format. The experiments were performed on a Nvidia A40 GPU under PyTorch v1.10.1, Python 3.9.7, and CUDA 11.4, with $batchsize=256$, $learning rate=0.0002$ with Adam optimizers and hyperparameter settings as $\lambda_{adv}=1,\lambda_{con}=50,\lambda_{adcon}=15,\lambda_{lat}=5$. To solely test the proposed method, no data augmentation is performed apart from image normalization. Notably, this section presented the common one-class anomaly detection performance, which means each model is trained on one single class, and tested against all other classes.

% Experiments for multi-class anomaly detection is in the appendix \Cref{sec:mc}.

\begin{table*}
    \scriptsize
    \begin{adjustbox}{width=.9\hsize,center}
    \begin{tabular}{c r c c c c c c c c c c c c }
    \toprule
    Dataset &  Method & 0 & 1 & 2 & 3 & 4 & 5 & 6 & 7 & 8 & 9 & avg & SD   \\[-1ex]
    ------ & ------ & --- & --- & --- & --- & --- & --- & --- & --- & --- & --- & --- & --- \\[-1ex]
    \multirow{5}*{MNIST}
        & AnoGAN & 99.0 & 99.8 & 88.8 & 91.3 & 94.4 & 91.2 & 92.5 & 96.4 & 88.3 & 95.8 & 93.7 & 4.00 \\
        & OCGAN & 99.8 & 99.9 & 94.2 & 96.3 & 97.5 & 98.0 & 99.1 & 98.1 & 93.9 & 98.1 & 97.5 & 2.10\\
        & GeoTrans & 98.2 & 91.6 & 99.4 & 99.0 & 99.1 & {\bfseries 99.6} & {\bfseries 99.9} & 96.3 & {\bfseries 97.2} & 99.2 & 98.0 & 2.50\\
        & ARNet & 98.6 & 99.9 & 99.0 & {\bfseries 99.1} & 98.1 & 98.1 & 99.7 & 99.0 & 93.6 & 97.8 & 98.3 & 1.78 \\
        & *ADGAN & 92.9 & 99.9 & 80.0 & 65.0 & 84.6 & 82.5 & 68.1 & 85.3 & 77.2 & 74.4 & 81.0 & 10.07 \\[-1ex]
        & ------ & --- & --- & --- & --- & --- & --- & --- & --- & --- & --- & --- & --- \\[-1ex]
        & Ours\tablefootnote{\label{tablefn} trained with naive encoder-decoder than UNet++.} & {\bfseries 100.} & {\bfseries 100.} & {\bfseries 99.0} & 98.6 & {\bfseries 99.5} & 97.2 & 99.6 & {\bfseries 99.8} & 95.8 & {\bfseries 99.2} & {\bfseries 99.1} & {\bfseries 0.86} \\
        % & Ours (UNet) & 76.0 & 58.6 & 81.3 & 93.3 & 41.9 & 46.8 & 98.9 & 45.2 &  89.1 &  42.8 &  67.4 &  21.6 \\
        % & Ours (UNet++) & 95.5 & 100. & 85.4 & 78.5 & 91.4 & 92.5 & 73.6 &  90.2 &  89.6 &  89.4 &  89.4 &  7.46 \\
        \cmidrule{2-14}
    \multirow{4}*{\makecell{ Fashion- \\ MNIST }}
        &  Method & 0 & 1 & 2 & 3 & 4 & 5 & 6 & 7 & 8 & 9 & avg & SD   \\[-1ex]
        & ------ & --- & --- & --- & --- & --- & --- & --- & --- & --- & --- & --- & --- \\[-1ex]
        & GeoTrans & 99.4 & 97.6 & 91.1 & 89.9 & 92.1 & 93.4 & 83.3 & 98.9 & 90.8 & 99.2 & 93.5 & 5.22 \\
        & ARNet & 92.7 & 99.3 & 89.1 & 93.6 & 90.8 & 93.1 & 85.0 & 98.4 & 97.8 & 98.4 & 93.9 & 4.70 \\
        & *ADGAN & 99.3 & {\bfseries 100.} & {\bfseries 100.} & 97.5 & {\bfseries 100.} & {\bfseries 100.} & 97.7 & {\bfseries 100.} & {\bfseries 100.} & {\bfseries 100.} & 99.5 & 0.95 \\[-1ex]
        & ------ & --- & --- & --- & --- & --- & --- & --- & --- & --- & --- & --- & --- \\[-1ex]
        % & Ours (AE) & 98.2 & 100. & 99.1 & 99.7 & 99.0 & 100. & 97.7 &  100. &  99.1 &  100. &  99.2 &  0.75 \\
        % & Ours (UNet) & {\bfseries 95.5} & {\bfseries 99.8} & {\bfseries 99.6} & {\bfseries 97.1} & {\bfseries 99.9} & {\bfseries 99.9} & {\bfseries 93.2} & {\bfseries 100.} & {\bfseries 99.4} & {\bfseries 99.0} & {\bfseries 98.3} & {\bfseries 2.21}  \\
        & Ours & {\bfseries 99.9} & {\bfseries 100.} & {\bfseries 100.} &  {\bfseries 99.4} & {\bfseries 100.} & {\bfseries 100.} & {\bfseries 100.} & {\bfseries 100.} & {\bfseries 100.} & {\bfseries 100.} & {\bfseries 99.9} & {\bfseries 0.18} \\
        \cmidrule{2-14}
    \multirow{7}*{\makecell{ CIFAR10 }}
        &  Method & 0 & 1 & 2 & 3 & 4 & 5 & 6 & 7 & 8 & 9 & avg & SD  \\[-1ex]
        & ------ & --- & --- & --- & --- & --- & --- & --- & --- & --- & --- & --- & --- \\[-1ex]
        & AnoGAN & 64.0 & 56.5 & 64.8 & 52.8 & 67.0 & 59.2 & 62.5 & 57.6 & 72.3 & 58.2 & 61.2 & 5.68 \\
        & OCGAN & 75.7 & 53.0 & 64.0 & 62.0 & 72.3 & 62.0 & 72.3 & 57.5 & 82.0 & 55.4 & 65.6 & 9.52 \\
        & GeoTrans & 74.7 & 95.7 & 78.1 & 72.4 & 87.8 & 87.8 & 83.4 & {95.5} & 93.3 & 91.3 & 86.0 & 8.52 \\
        & ARNet & 78.5 & 89.8 & 86.1 & 77.4 & 90.5 & 84.5 & 89.2 & 92.9 & 92.0 & 85.5 & 86.6 & 5.35 \\
        & ADGAN & {99.9} & 91.5 & 90.9 & 91.7 & {99.5} & 86.1 & 99.7 & 86.3 & 99.9 & {93.1} & {93.8} & {5.25} \\[-1ex]
        & ------ & --- & --- & --- & --- & --- & --- & --- & --- & --- & --- & --- & --- \\[-1ex]
        % & Ours (AE) & 79.9 & 59.1 & 64.9 & 60.1 &  68.9 &  69.8 &  88.2 &  61.3 &  83.0 &  74.3 &  71.0 &  9.63 \\
        % & Ours (UNet) & {\bfseries 99.9} & {\bfseries 88.3} & {\bfseries 89.8} & {\bfseries 94.1} & {99.3} & {\bfseries 87.7} & {\bfseries 99.8} & 73.8 & {\bfseries 99.9} & 92.3 & 92.5 & 7.81 \\
        & Ours & {\bfseries 99.9} & {\bfseries 99.5} & {\bfseries 98.8} & {\bfseries 98.4} & {\bfseries 99.7} & {\bfseries 94.7} & {\bfseries 100.} & {\bfseries 93.2} & {\bfseries 99.9} & {\bfseries 98.8} & {\bfseries 98.3} & {\bfseries 2.26} \\
    \midrule
    \end{tabular}
    \end{adjustbox}

    \begin{adjustbox}{width=.9\hsize,center}
    \begin{tabular}{c r c c c c c c c c c c c}
    \multirow{13}*{CIFAR100}
        &  Method & 0 & 1 & 2 & 3 & 4 & 5 & 6 & 7 & 8 & 9 & 10  \\[-1ex]
        & ------ & --- & --- & --- & --- & --- & --- & --- & --- & --- & --- & --- \\[-1ex]
        & DAGMM & 43.4 & 49.5 & 66.1 & 52.6 & 56.9 & 52.4 & 55.0 & 52.8 & 53.2 & 42.5 & 52.7 \\
        & GeoTrans & 74.7 & 68.5 & 74.0 & 81.0 & 78.4 & 59.1 & 81.8 & 65.0 & 85.5 & 90.6 & 87.6 \\
        & ARNet & 77.5 & 70.0 & 62.4 & 76.2 & 77.7 & 64.0 & 86.9 & 65.6 & 82.7 & 90.2 & 85.9 \\
        & ADGAN & 89.1 & 69.4 & 96.5 & 90.6 & 89.5 & 83.7 & 78.9 & 57.2 & {90.2}& 94.8 & 83.0\\[-1ex]
        & ------ & --- & --- & --- & --- & --- & --- & --- & --- & --- & --- & --- \\[-1ex]
        % & Ours (UNet) & {\bfseries 99.9} & {\bfseries 99.4} & {\bfseries 99.7} & {\bfseries 99.3} & {\bfseries 98.9} & {\bfseries 97.5} & {\bfseries 99.8} & {\bfseries 96.0} & {\bfseries 99.4} & {\bfseries 99.3} & {\bfseries 99.3} \\
        & Ours & {\bfseries 99.9} & {\bfseries 99.6} & {\bfseries 99.8} & {\bfseries 99.8} & {\bfseries 99.9} & {\bfseries 99.7} & {\bfseries 99.9} & {\bfseries 99.9} & {\bfseries 99.4} & {\bfseries 99.8} & {\bfseries 99.9} \\
        \cmidrule{2-13}
        &  Method & 11 & 12 & 13 & 14 & 15 & 16 & 17 & 18 & 19 & avg & SD  \\[-1ex]
        & ------ & --- & --- & --- & --- & --- & --- & --- & --- & --- & --- & --- \\[-1ex]
        % & AE & 62.1 & 59.6 & 49.8 & 48.1 & 56.4 & 57.6 & 47.2 & 47.1 & 41.5 & 52.4 & 8.11\\
        & DAGMM & 46.4 & 42.7 & 45.4 & 57.2 & 48.8 & 54.4 & 36.4 & 52.4 & 50.3 & 50.5 & {\bfseries 6.55}\\
        & GeoTrans & 83.9 & 83.2 & 58.0 & 92.1 & 68.3 & 73.5 & 93.8 & {\bfseries 90.7} & 85.0 & 78.7 & 10.76 \\
        & ARNet & 83.5 & 84.6 & 67.6 & 84.2 & {\bfseries 74.1} & 80.3 & 91.0 & 85.3 & 85.4 & 78.8 & 8.82 \\
        & ADGAN & 86.8 & 96.2 & 71.7 & {94.1} & 71.4 & 79.1 & 94.7 & 80.7 & 72.4 & 83.5 & 10.53  \\[-1ex]
        & ------ & --- & --- & --- & --- & --- & --- & --- & --- & --- & --- & --- \\[-1ex]
        % & Ours (UNet) & {\bfseries 96.3} & {\bfseries 99.8} & {\bfseries 74.7} & 78.9 & 62.6 & {\bfseries 95.6} & {\bfseries 95.5} & 66.8 & {\bfseries 96.9} & {\bfseries 92.8} & 11.5 \\
        & Ours & {\bfseries 99.1} & {\bfseries 99.9} & {\bfseries 86.4} & {\bfseries 96.9} & 61.7 & {\bfseries 99.0} & {\bfseries 99.8} & 86.5 & {\bfseries 99.7} & {\bfseries 96.3} & 8.88 \\
    \bottomrule
    \end{tabular}
    \end{adjustbox}
    \caption{One-class semi-supervised anomaly detection benchmark performances. We reported the average AUC in \% that is computed over 3 runs. Results of AnoGAN \cite{anogan-1703.05921}, OCGAN \cite{Perera2019}, GeoTrans \cite{GeoTrans1805.10917}, ADGAN \cite{Cheng2020}, and DAGMM \cite{Zong2018DeepAG} are borrowed from \cite{Cheng2020}, while the results of ARNet are borrowed from \cite{Ye2022}. Results that marked with * are produced by us. Bold numbers represent the best results.}
    \label{tab:oc_ssup}
\end{table*}

\subsubsection{Semi-supervised Anomaly Detection} We compared our method with several influential and performant reconstruction-based semi-supervised deep anomaly detection models, including AnoGAN \cite{anogan-1703.05921}, OCGAN \cite{Perera2019}, GeoTrans \cite{GeoTrans1805.10917}, ARNet \cite{Ye2022}, and ADGAN \cite{Cheng2020}. \Cref{tab:oc_ssup} illustrated the results across those models and the benchmarking datasets. The 0-9 headers represent the corresponding 10 classes for MNIST, Fashion-MNIST and CIFAR10 datasets, while the 0-19 headers represent the 20 coarse classes of CIFAR100. "SD" means standard deviation among classes. Detailed index-class mapping refers to \Cref{sec:label_tag}. As shown in \Cref{tab:oc_ssup}, the average AUC show significant improvement against other methods with 0.8\% better in MNIST, 6.0\% better in Fashion MNIST, 4.5\% better in CIFAR10, and 12.8\% better in CIFAR100.

\begin{table*}
    \scriptsize
    \begin{adjustbox}{width=.9\hsize,center}
    \begin{tabular}{c c c c c c c c c c c c c c c }
    \hline
    Dataset & $\gamma$ & \makecell{supervised \\ classifier} & SS-DGM & \makecell{SSAD \\ Hybrid} & Deep SAD & TLSAD  & ESAD & Ours  \\
    \toprule
    \multirow{5}*{MNIST\tablefootnotemark{tablefn}}
        & .00 & - &      -       & 96.3$\pm$2.5 & 92.8$\pm$4.9 &  -   & 98.5$\pm$1.3 & {\bfseries 99.1}$\pm${\bfseries 0.86} \\
        & .01 & 83.6$\pm$8.2 & 89.9$\pm$9.2 & 96.8$\pm$2.3 & 96.4$\pm$2.7 & 94.1 & 99.2$\pm$0.7 & {\bfseries 99.4}$\pm${\bfseries 0.85} \\
        & .05 & 90.3$\pm$4.6 & 92.2$\pm$5.6 & 97.4$\pm$2.0 & 96.7$\pm$2.3 & 96.9 & 99.4$\pm$0.3 & {\bfseries 99.9}$\pm${\bfseries 0.29} \\
        & .10 & 93.9$\pm$2.8 & 91.6$\pm$5.5 & 97.6$\pm$1.7 & 96.9$\pm$2.3 & 97.7 & 99.5$\pm$0.4 & {\bfseries 100.}$\pm${\bfseries 0.04} \\
        & .20 & 96.9$\pm$1.7 & 91.2$\pm$5.6 & 97.8$\pm$1.5 & 96.9$\pm$2.4 & 98.3 & 99.6$\pm$0.3 & {\bfseries 100.}$\pm${\bfseries 0.02} \\
    \midrule
    \multirow{5}*{\makecell{Fashion-\\MNIST}}
        & .00 &      -        &       -       & 91.2$\pm$4.7 & 89.2$\pm$6.2 & -& 94.0$\pm$4.5 & {\bfseries 99.9}$\pm${\bfseries 0.18} \\
        & .01 & 74.4$\pm$13.6 & 65.1$\pm$16.3 & 89.4$\pm$6.0 & 90.0$\pm$6.4 & 88.4 & 95.3$\pm$4.2 & {\bfseries 100.}$\pm${\bfseries 0.00} \\
        & .05 & 76.8$\pm$13.2 & 71.4$\pm$12.7 & 90.5$\pm$5.9 & 90.5$\pm$6.5 & 91.4 & 95.6$\pm$4.1 & {\bfseries 100.}$\pm${\bfseries 0.00} \\
        & .10 & 79.0$\pm$12.3 & 72.9$\pm$12.2 & 91.0$\pm$5.6 & 91.3$\pm$6.0 & 92.0 & 95.8$\pm$4.0 & {\bfseries 100.}$\pm${\bfseries 0.00}  \\
        & .20 & 81.4$\pm$12.0 & 74.7$\pm$13.5 & 89.7$\pm$6.6 & 91.0$\pm$5.5 & 93.2 & 95.9$\pm$4.0 & {\bfseries 100.}$\pm${\bfseries 0.00}  \\
    \midrule
    % UNet version
    % \multirow{5}*{CIFAR10}
    %     & .00 &        -      &           -    & 63.8$\pm$9.0 & 60.9$\pm$9.4 & -   & 78.8$\pm${\bfseries 6.5} & {\bfseries 92.5}$\pm$7.81 \\
    %     & .01 & 55.6$\pm$5.0 & 49.7$\pm$1.7 & 70.5$\pm$8.30 & 72.6$\pm$7.4 & 74.4 & 83.7$\pm$6.4 & {\bfseries 94.8}$\pm${\bfseries 6.20} \\
    %     & .05 & 63.5$\pm$8.0 & 50.8$\pm$4.7 & 73.3$\pm$8.4 & 77.9$\pm$7.2 & 80.0 & 86.9$\pm$6.8  & {\bfseries 98.3}$\pm${\bfseries 2.38} \\
    %     & .10 & 67.7$\pm$9.6 & 52.0$\pm$5.5 & 74.0$\pm$8.1 & 79.8$\pm$7.1 & 84.8 & 87.8$\pm$6.7  & {\bfseries 99.3}$\pm${\bfseries 1.15} \\
    %     & .20 & 80.5$\pm$5.9 & 53.2$\pm$6.7 & 74.5$\pm$8.0 & 81.9$\pm$7.0 & 86.3 & 88.5$\pm$6.9  & {\bfseries 99.9}$\pm${\bfseries 0.15} \\
    \multirow{5}*{CIFAR10}
        & .00 &        -      &           -    & 63.8$\pm$9.0 & 60.9$\pm$9.4 & -   & 78.8$\pm$6.5 & {\bfseries 98.3}$\pm${\bfseries 2.26} \\
        & .01 & 55.6$\pm$5.0 & 49.7$\pm$1.7 & 70.5$\pm$8.30 & 72.6$\pm$7.4 & 74.4 & 83.7$\pm$6.4 & {\bfseries 98.6}$\pm${\bfseries 1.27} \\
        & .05 & 63.5$\pm$8.0 & 50.8$\pm$4.7 & 73.3$\pm$8.4 & 77.9$\pm$7.2 & 80.0 & 86.9$\pm$6.8  & {\bfseries 99.8}$\pm${\bfseries 0.25} \\
        & .10 & 67.7$\pm$9.6 & 52.0$\pm$5.5 & 74.0$\pm$8.1 & 79.8$\pm$7.1 & 84.8 & 87.8$\pm$6.7  & {\bfseries 99.9}$\pm${\bfseries 0.03} \\
        & .20 & 80.5$\pm$5.9 & 53.2$\pm$6.7 & 74.5$\pm$8.0 & 81.9$\pm$7.0 & 86.3 & 88.5$\pm$6.9  & {\bfseries 100.}$\pm${\bfseries 0.00} \\
    \bottomrule  
    % \multirow{5}*{CIFAR100}
    %     & .00 & & & &  & 92.2$\pm$11.2 \\
    %     & .01 & & & &  & 97.1$\pm$5.99 \\
    %     & .05 & & & &  & 98.4$\pm$3.99 \\
    %     & .10 & & & &  &  \\
    %     & .20 & & & &  & 99.7$\pm$0.92 \\
    % \hline
    \end{tabular}
    
    \end{adjustbox}
    \caption{\small One-class anomaly detection benchmark performances with increased supervision. Results of supervised classifier, SS-DGM, SSAD Hybrid, and Deep SAD are borrowed from \cite{ruff2020deep}, while the results of TLSAD, and ESAD are borrowed from \cite{ESAD2020}. Bold numbers represent the best results.}
    \label{tab:oc_sup}
\end{table*}

\subsubsection{Anomaly Detection with Limited Supervision} \Cref{tab:oc_sup} shows the benchmarking results when limited anomaly supervision brought in. Here, we compared with several supervised anomaly detection methods like Deep SAD \cite{ruff2020deep}, TLSAD \cite{Feng2021}, and ESAD \cite{ESAD2020}. In this experiment, we gradually increased the involvement percentage $\gamma$ of anomaly data. Here, anomaly percentage $\gamma$ represented $\gamma=\frac{a}{n+a}$ where $n$ and $a$ are the number of normal and abnormal images respectively. Also, an average of three runs is recorded for each class and each dataset to mitigate the selection bias whilst data sampling. As we observed, our method can significantly improve the model performances with $\gamma$ increased. Specifically, our method showed significant improvement with around 0.2\% better in MNIST, 4\% better in Fashion-MNIST, and 15\% better in CIFAR10 at each $\gamma$ scale. Model performances turn to be robust with only 5\% anomaly data affixed.

\subsection{Qualitative Analysis}

\begin{figure*}
    \centering
    \def\imagetop#1{\raisebox{-.4\height+\baselineskip}{#1}}
    \begin{tabularx}{\textwidth}{cccc}
        \makecell{Source}
            & \imagetop{\includegraphics[width=.252\textwidth]{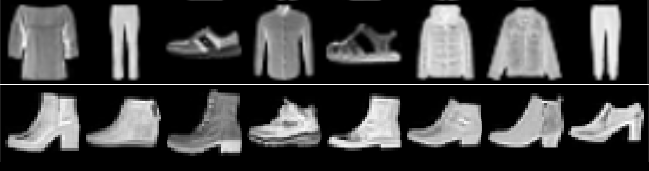}}
            & \imagetop{\includegraphics[width=.252\textwidth]{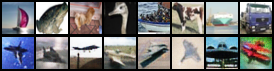}}
            & \imagetop{\includegraphics[width=.252\textwidth]{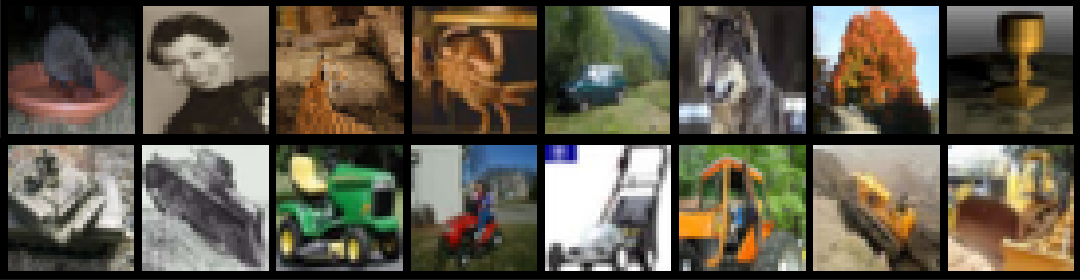}} \\
        \makecell{GANomaly}
            & \imagetop{\includegraphics[width=.252\textwidth]{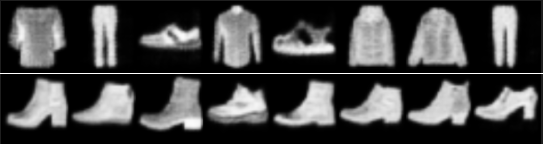}}
            & \imagetop{\includegraphics[width=.252\textwidth]{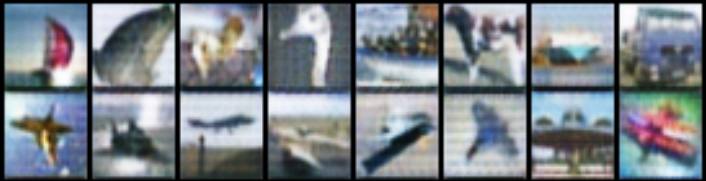}}
            & \imagetop{\includegraphics[width=.252\textwidth]{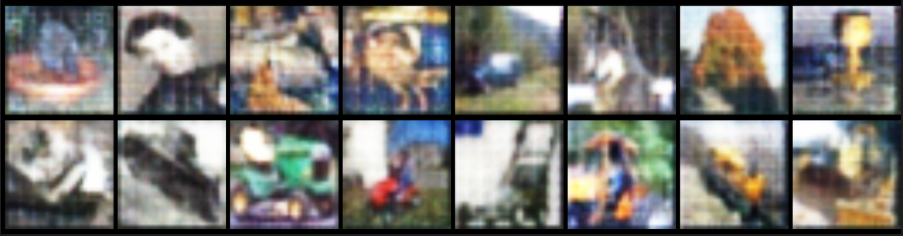}} \\
        \makecell{Ours \\ (unsupervised)}
            & \imagetop{\includegraphics[width=.252\textwidth]{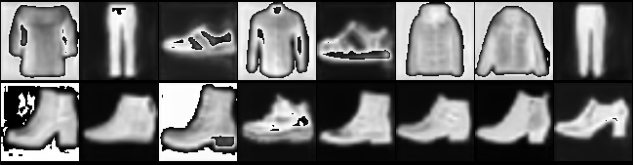}}
            & \imagetop{\includegraphics[width=.252\textwidth]{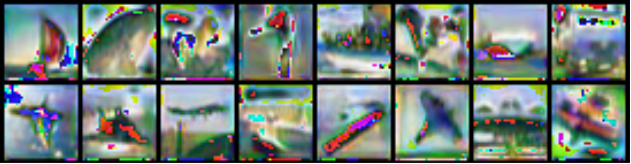}}
            & \imagetop{\includegraphics[width=.252\textwidth]{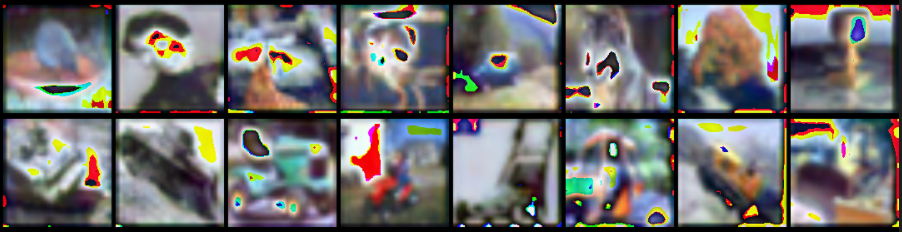}} \\
        \makecell{Ours \\ (supervised)}
            & \imagetop{\includegraphics[width=.252\textwidth]{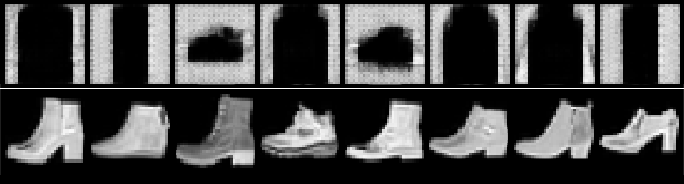}}
            & \imagetop{\includegraphics[width=.252\textwidth]{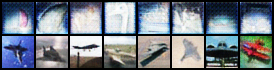}}
            & \imagetop{\includegraphics[width=.252\textwidth]{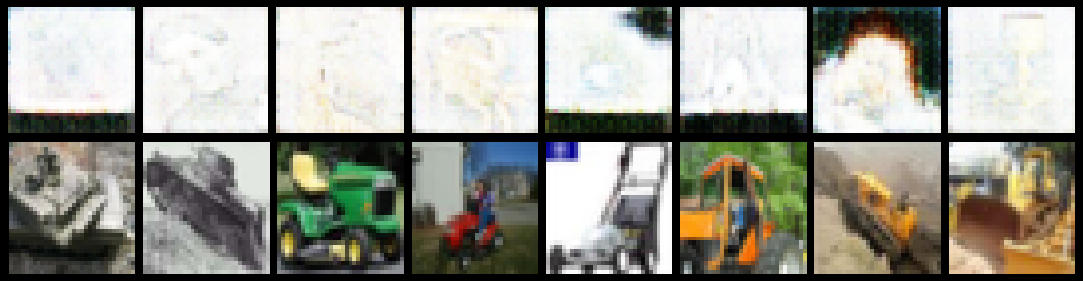}} \\
        & Fashin-MNIST & CIFAR10 & CIFAR100
    \end{tabularx}
    
    % \begin{minipage}[b]{.13\textwidth}
    %     % {\tiny A: non-boot\\N: boot \\}
    %     % \vspace{.2em}
    %     % {\tiny A: non-plane\\N: plane \\}
    %     % \vspace{.2em}
    %     % {\tiny A: non-vehicle\\N: vehicle \\}
    %     % \vspace{-.6em}
    %     {\tiny Source images \vspace{1.8em} \\ \\}
    %     {\tiny GANomaly \vspace{1.em} \\}
    %     {\tiny Ours \\ (Unsupervised) \vspace{-1em} \\}
    %     {\tiny Ours \\ (supervised) \vspace{-1em} \\}
        
    % \end{minipage}
    % \subfigure[Fashion-MNIST]{
    % \begin{minipage}[b]{.252\textwidth}
    %     \includegraphics[width=\textwidth]{reals-fmnist.png}
    %     \includegraphics[width=\textwidth]{fakes-ganomaly-fmnist.png}
    %     \includegraphics[width=\textwidth]{fakes-fmnist_unsup.png}
    %     \includegraphics[width=\textwidth]{fakes-fmnist.png}
    % \end{minipage}
    % }
    % \subfigure[CIFAR10]{
    % \begin{minipage}[b]{.26\textwidth}
    %     \includegraphics[width=\textwidth]{cifar10-reals.png}
    %     \includegraphics[width=\textwidth]{fakes-ganomaly.png}
    %     \includegraphics[width=\textwidth]{fakes-cifar10_unsup.png}
    %     \includegraphics[width=\textwidth]{cifar10-fakes.png}
    % \end{minipage}
    % }
    % \subfigure[CIFAR100]{
    % \begin{minipage}[b]{.258\textwidth}
    %     \includegraphics[width=\textwidth]{reals-cifar100.png}
    %     \includegraphics[width=\textwidth]{fakes-ganomaly-cifar100.png}
    %     \includegraphics[width=\textwidth]{fakes-cifar100_unsup.png}
    %     \includegraphics[width=\textwidth]{fakes-cifar100.png}
    % \end{minipage}
    % }
    \caption{Reconstruction results for Fashion-MNIST, CIFAR10, and CIFAR100. For each dataset, 16 images are randomly chosen for visualization, including 8 anomaly images (upper) and 8 normal images (lower). Here, "ankle boot", "airplane", and "vehicle 2" are the anomalous classes for Fashion-MNIST, CIFAR10, and CIFAR100 respectively. The supervised models used here are trained with 10\% anomaly data.}
    \label{fig:rec}
\end{figure*}

This section presents a qualitative analysis of our proposed AGAD method regarding the reconstruction of anomalous and non-anomalous data. This subsection compares our reconstruction quality against GANomaly \cite{Akcay2019-jx}. As shown in \Cref{fig:rec}, GANomaly reconstructs both normality and abnormality data with minor reconstruction error, while our method is designed to discriminate the reconstruction of normality and abnormality data for better anomaly detection performance. Particularly, our supervised approach identifies the expected anomalous patterns that reconstructed well for normal examples while messed up reconstructions for anomalous data.

\section{Experiments}\label{app:realworld}

We further evaluate our method on real-world medical datasets. Different from the benchmark datasets, real-world anomaly features might be less significant. Images have been resized to 128x128 for retaining more information.

\subsection{Real-world Datasets}
We experiment on real-world medical datasets including X-ray, Brain MRI, histopathology and retinal OCT images.

\begin{itemize}[noitemsep,leftmargin=*]

    \item{\bfseries Alzheimer's Dataset} \cite{AlzheimerData} contains 6,412 brain magnetic resonance imaging (MRI) scans with 4 categories (normal, very mild demented, mild demented, and moderate demented). Training set contains 2,560 normal, 1,792 very mild demented, 717 mild demented, and 52 moderate demented images. Testing set contains 640 normal, 448 very mild demented, 179 mild demented, and 12 moderate demented images.

    \item{\bfseries ChestXray} \cite{Kermany2018-eu} contains 5,863 X-Ray images and 2 categories (pneumonia/normal), with a training set of 1,341 normal and 3,875 pneumonia images, and a testing set (from the merge of original test and val sets) of 242 normal and 398 pneumonia images.

    \item{\bfseries Lung Histopathology} is the lung histopathology subset of LC25000 dataset \cite{Lungcolondataset} that contains 5,000 lung adenocarcinoma (ACA) images, 5,000 lung squamous cell carcinomas (SCC) images and 5,000 benign lung images. As the training/testing sets are not splitted, we selected the last 500 images in each class as testing set alphabetically.
    
    \item{\bfseries Retinal OCT} \cite{Kermany2018-eu} consists 84,495 retinal optical coherence tomography (OCT) images with 4 categories of choroidal neovascularization (CNV), diabetic macular edema (DME), drusen and normal. The training set contains 37,205 CNV, 11,348 DME, 8,616 drusen, 26,315 normal images, while the testing set (from the merge of original test and val sets) contains 250 images for each category.

    % \item{\bfseries Colon Histopathology} is the colon histopathology subset of LC25000 dataset \cite{Lungcolondataset} that contains 5,000 colon adenocarcinoma (ACA) images and 5,000 benign colonic images. As the training/testing sets are not splitted, we selected the last 500 images in each class as testing set alphabetically.

\end{itemize}

\begin{figure*}
    \centering
    {
        \begin{minipage}[b]{.4\textwidth}
            \includegraphics[width=.322\textwidth]{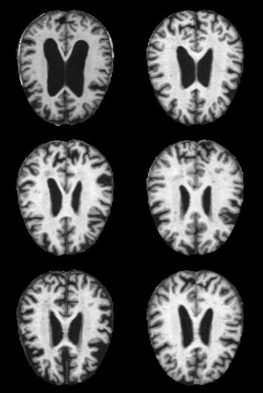}
            \includegraphics[width=.322\textwidth]{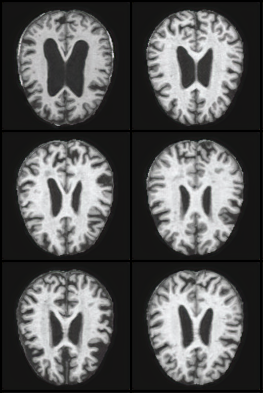}
            \includegraphics[width=.322\textwidth]{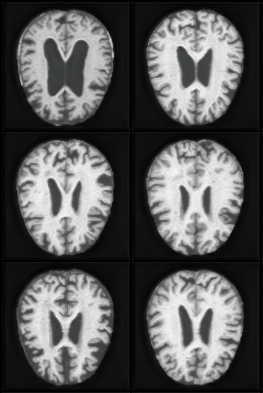}
        \end{minipage}
        \hspace{0.04\textwidth}
        \begin{minipage}[b]{.4\textwidth}
            \includegraphics[width=.322\textwidth]{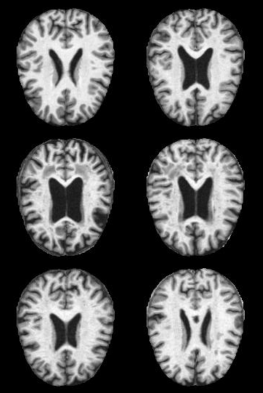}
            \includegraphics[width=.322\textwidth]{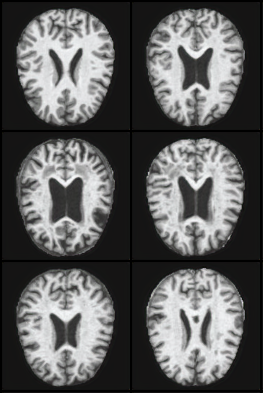}
            \includegraphics[width=.322\textwidth]{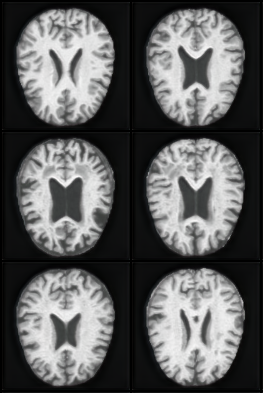}
        \end{minipage}
    }
    {
        \begin{minipage}[b]{.4\textwidth}
            \includegraphics[width=.322\textwidth]{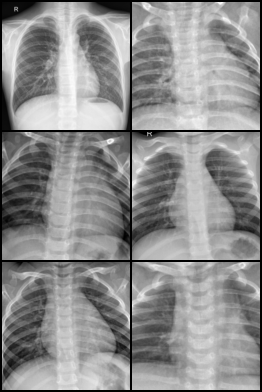}
            \includegraphics[width=.322\textwidth]{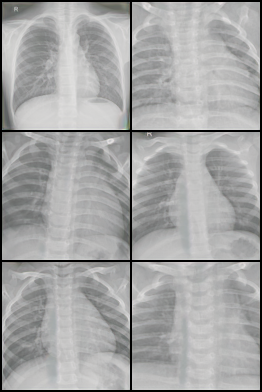}
            \includegraphics[width=.322\textwidth]{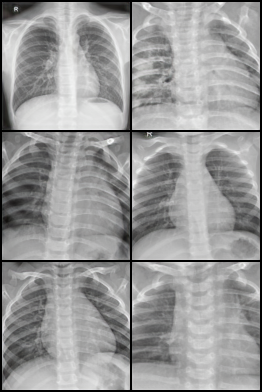}
        \end{minipage}
        \hspace{0.04\textwidth}
        \begin{minipage}[b]{.4\textwidth}
            \includegraphics[width=.322\textwidth]{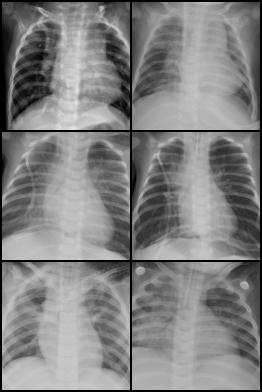}
            \includegraphics[width=.322\textwidth]{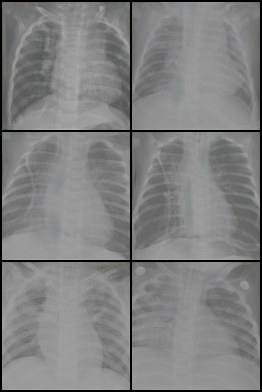}
            \includegraphics[width=.322\textwidth]{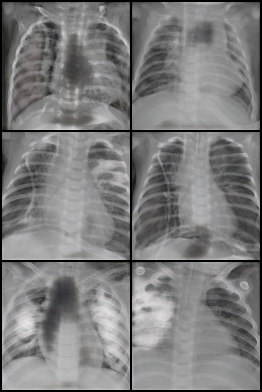}
        \end{minipage}
    }
    {
        \begin{minipage}[b]{.4\textwidth}
            \includegraphics[width=.322\textwidth]{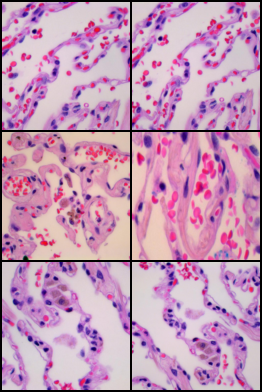}
            \includegraphics[width=.322\textwidth]{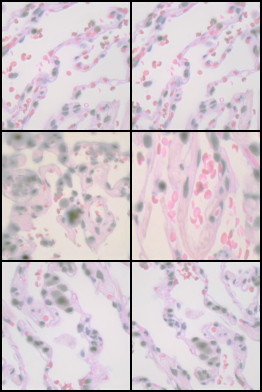}
            \includegraphics[width=.322\textwidth]{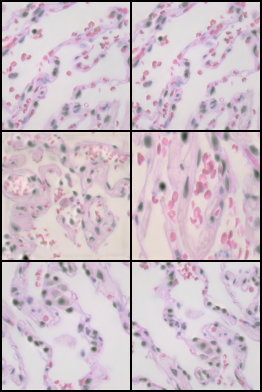}
        \end{minipage}
        \hspace{0.04\textwidth}
        \begin{minipage}[b]{.4\textwidth}
            \includegraphics[width=.322\textwidth]{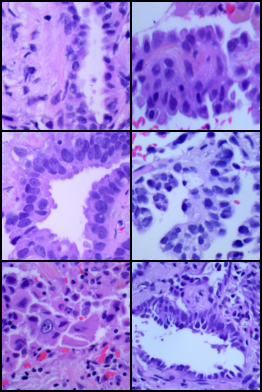}
            \includegraphics[width=.322\textwidth]{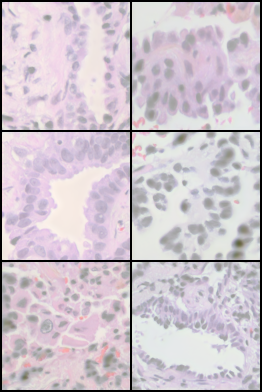}
            \includegraphics[width=.322\textwidth]{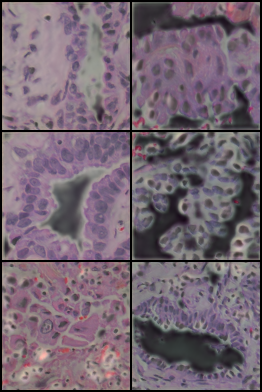}
        \end{minipage}
    }
    {
        \begin{minipage}[b]{.4\textwidth}
            \includegraphics[width=.322\textwidth]{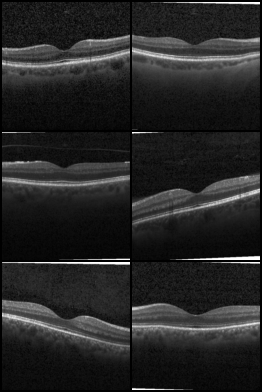}
            \includegraphics[width=.322\textwidth]{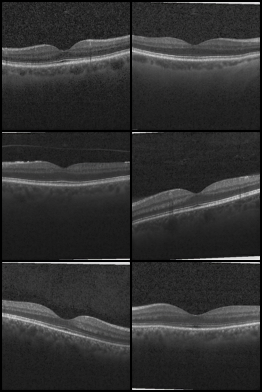}
            \includegraphics[width=.322\textwidth]{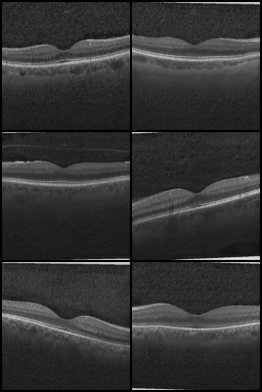}
        \end{minipage}
        \hspace{0.04\textwidth}
        \begin{minipage}[b]{.4\textwidth}
            \includegraphics[width=.322\textwidth]{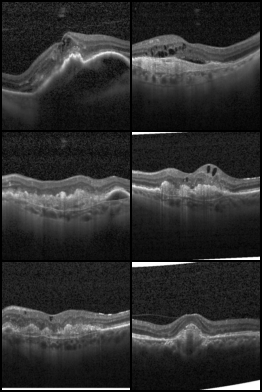}
            \includegraphics[width=.322\textwidth]{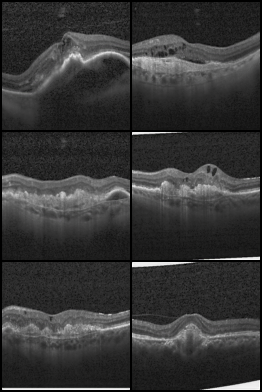}
            \includegraphics[width=.322\textwidth]{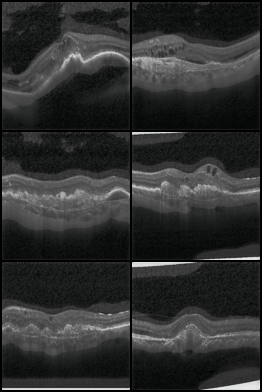}
        \end{minipage}
    }
    {\footnotesize
    \begin{tabular}{c c c c c c c}
       ~~~~~\makecell{Normal\\images}~~~~~  & ~~~~~\makecell{Ours\\(unsupervised)}~~ & ~~~~~\makecell{Ours\\(supervised)}~~ & ~~~~~~ & ~~~\makecell{Abnormal\\images}~~  & ~~~~~\makecell{Ours\\(unsupervised)} & ~~~~~~~~\makecell{Ours\\(supervised)}  \\
    %   \\
    %     \multicolumn{3}{c}{Normal} & & \multicolumn{3}{c}{Abnormal} \\
    \end{tabular}
    }
    \caption{Reconstruction results of the real-world datasets. From top to bottom, are the brain MRI, chest X-ray, lung histopathology, and retinal OCT data respectively. We randomly selected 6 normal images and 6 abnormal images for each dataset. Supervised models used for each datasets were trained with 10\% anomaly.}
    \label{fig:app_vis}
\end{figure*}

\begin{table*}
    \centering
    % \begin{adjustbox}{width=\linewidth,center}
    \begin{tabular}{c c c c c c c c c c c c c}
    \toprule
    % Dataset & \multicolumn{10}{c}{Anomaly percentage $\gamma$}   \\
    $\gamma$  &  \multicolumn{2}{c}{~~.00~~}  & \multicolumn{2}{c}{~~.01~~} & \multicolumn{2}{c}{~~.03~~} & \multicolumn{2}{c}{~~.05~~} & \multicolumn{2}{c}{~~.10~~} \\
    Adversarial Context & ~~w~ & ~w/o~ & ~~w~ & ~w/o~ & ~~w~ & ~w/o~ & ~~w~ & ~w/o~ & ~~w~ & ~w/o~ \\
    \midrule
    Alzheimer's               & {\bfseries 99.1} & 93.9 & {\bfseries 99.0} & 94.7 & {\bfseries 99.3} & 89.7 & {\bfseries 99.8} & 99.5 & 99.7 & {\bfseries 100.} \\
    *ChestXray                 & {\bfseries 72.6} & 68.8 & - & - & {\bfseries 95.5} & 73.4 & {\bfseries 96.3} & 89.8  & {\bfseries 98.9} & 91.4 \\
    Lung Histopathology        & {\bfseries 82.4} & 80.0 & {\bfseries 91.4} & 91.0 & {\bfseries 94.0} & 89.0 & 96.6 & {\bfseries 97.0} & {\bfseries 97.7} & 92.3 \\
    Retinal OCT                & {\bfseries 94.5} & 90.9 & {\bfseries 99.7} & 93.0  & {\bfseries 99.2} & 75.8 & {\bfseries 99.5} & 72.3 & {\bfseries 99.7} & 88.0 \\
    % Colon Histopathology    & 41.1 & 44.5 \\
    \bottomrule
    \end{tabular}
    % \end{adjustbox}
    \caption{One-class anomaly detection performances on medical datasets. We report the average AUC in \% that computed over 3 runs. Datasets marked with * got too few anomalies to experiment when $\gamma=0.01$. Bold numbers represent the best results.}
    \label{tab:my_label2}
\end{table*}

\subsubsection{Results}

\Cref{sec:benchmark} proved the effectiveness of our proposed method with a comprehensive benchmarking against several benchmark datasets and models. In this section, we further evaluated our model on several real-world medical datasets. The results demonstrated robustness over those datasets, especially with the proposed \textit{Contextual adversarial information}. Interestingly, network training would be failed without \textit{Contextual adversarial information} for Retinal OCT dataset when more supervision provided ($\gamma=0.03,0.05,0.10$). Due to the similarities between normal and abnormal data in Retinal OCT dataset, we suspect the similar normal and abnormal data confuses the network, resulting in failure. Therefore, we would refer the \textit{Contextual adversarial information} as a method that might accelerate learning on fine-grained anomaly features.

\subsubsection{Qualitative Analysis} In \Cref{fig:app_vis}, we further analysed the reconstruction results of those real-world datasets. The visual results got insignificant reconstruction errors for the Alzheimer's MRI data, while chest X-Ray and histopathology data obtained more meaningful results that mostly with corrupted the lung and edema areas. Though retinal OCT reached a good AUC (0.997), the corrupted reconstruction areas are less meaningful that mostly lie in background. In summary, we believe our method can produce reasonable reconstructions for anomaly detection tasks when detecting significant anomaly features in such highly structured medical imaging data.

\begin{figure*}
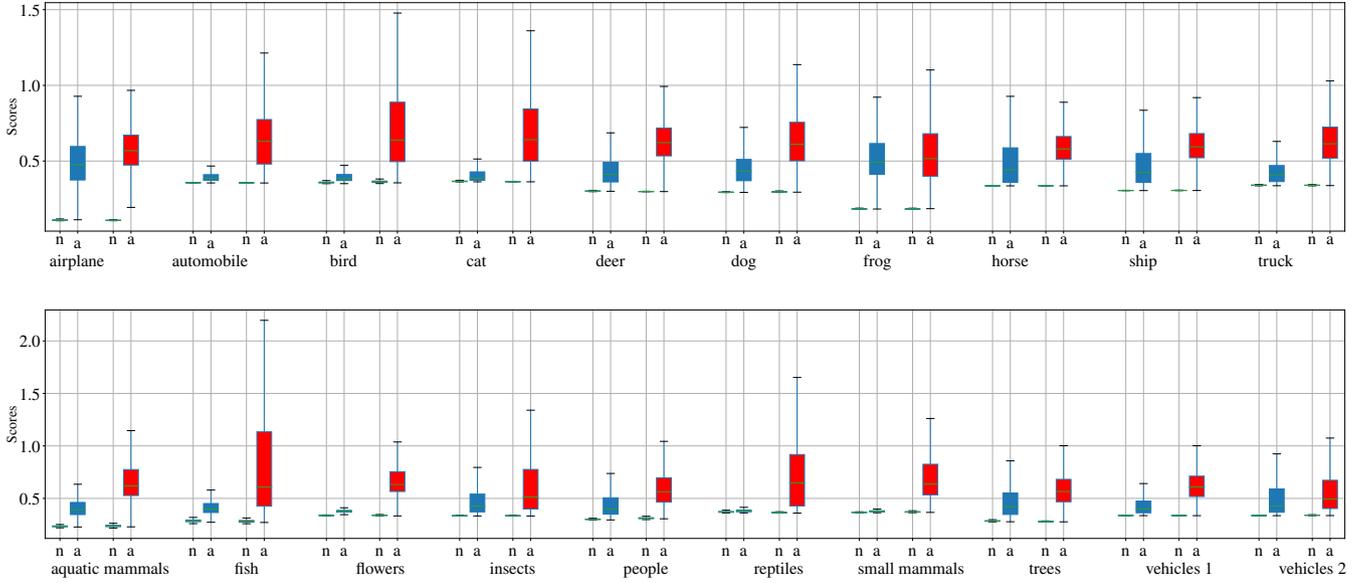

    \begin{adjustbox}{width=1.23\linewidth,center}
        \input{plot-cifar-2.pgf}
    \end{adjustbox}
    \begin{adjustbox}{width=1.23\linewidth,center}
        \input{plot-cifar100.pgf}
    \end{adjustbox}
    \caption{Class-wise box plot. Up: box plot for CIFAR10 dataset. Bottom: box plot for CIFAR100 dataset. Blue boxes represent the model trained with 5\% anomaly data, while red boxes represent the model trained with 20\% anomaly data. We randomly picked 10 classes out of the CIFAR100 dataset for demonstration purpose.}
    \label{fig:boxplot}
\end{figure*}

\section{Ablation Study}

This section performs a series of ablation studies to understand how our proposed method worked. Specifically, we explore the effectiveness of our proposed \textit{Contextual Adversarial Information}, as well as the value of the anomaly distributions. Then we further conduct statistical analysis to explore the effectiveness of the increasing amount of anomaly data. In this section, we use UNet \cite{UNet10.1007/978-3-319-24574-4_28} as the backbone generator, and all experiment settings are as same as in \Cref{sec:exp}.

\subsection{Contextual Adversarial Information}\label{sec:box}

This subsection demonstrates the effectiveness of the proposed \textit{contextual adversarial information}. As aforementioned, we expect to improve the data-efficiency
by generating pseudo-anomaly data when none or small number of anomaly data available. \Cref{tab:oc_compare} shows the our method improved model performances under limited anomaly data environment ($\gamma<0.05$), while the effect dimmed when sufficient supervision provided, indicating the proposed \textit{contextual adversarial information} mitigated the lack of anomaly data as expected.

% https://www.bmvc2021-virtualconference.com/assets/papers/0329.pdf

\begin{table}[pos=h]
    \scriptsize
    \centering
    % \begin{adjustbox}{width=\linewidth,center}
    \setlength{\tabcolsep}{2pt}
    \begin{tabular}{c l c c c c c c}
    \toprule
    Dataset &  C.A. & \multicolumn{5}{c}{Anomaly percentage $\gamma$}   \\
        &   & ~~~~.00~~~~ & ~~~~.01~~~~ & ~~~~.03~~~~ & ~~~~.05~~~~ & ~~~~.10~~~~ \\
    \midrule
    \multirow{2}*{\makecell{ F-MNIST }}
        & w/o & 98.2$\pm$2.37 & {99.9}±{0.03} & {\bfseries 100.}±{\bfseries 0.00} & {\bfseries 100.}±{\bfseries 0.00} & {\bfseries 100.}±{0.05} \\
        & w/  & {\bfseries 98.3}$\pm${\bfseries 2.21} & {\bfseries 100.}±{\bfseries 0.00} & {\bfseries 100.}±{\bfseries 0.00} & {\bfseries 100.}±{\bfseries 0.00} & {\bfseries 100.}±{\bfseries 0.00}  \\
        \cmidrule{2-7}
    \multirow{2}*{\makecell{ CIFAR10 }}
        & w/o & 88.9$\pm$10.43 & 93.8$\pm$5.30 & {98.0}±{2.74} & {98.8}±{2.16} & {99.3}±{1.09} \\
        & w/  & {\bfseries 92.6}$\pm${\bfseries 7.76} & {\bfseries 94.8}±{\bfseries 6.17} & {\bfseries 98.3}±{\bfseries 2.38} & {\bfseries 99.3}±{\bfseries 1.15} & {\bfseries 99.9}±{\bfseries 0.15} \\
        \cmidrule{2-7}
    \multirow{2}*{\makecell{ CIFAR100 }}
        & w/o & {90.4}$\pm${11.5} & {95.9}$\pm${7.76} & {96.7}$\pm${7.58} & {\bfseries 98.8}$\pm${\bfseries 3.09} & {99.5}$\pm${1.65}\\
        & w/  & {\bfseries 92.8}±{\bfseries 11.5} & {\bfseries 97.1}$\pm${\bfseries 6.20} & {\bfseries 97.8}$\pm${\bfseries 4.90} & {\bfseries 98.8}$\pm${3.63} & {\bfseries 99.7}$\pm${\bfseries 0.80}\ \\
    \bottomrule
    \end{tabular}
    % \end{adjustbox}
    \caption{Ablation study on \textit{contextual adversarial information}. We experimented on Fashion-MNIST, CIFAR10, and CIFAR100 datasets with different levels of anomaly involvement. Bold numbers represent the best results.}
    \label{tab:oc_compare}
\end{table}

\subsection{Batch Normalization for Anomaly data}\label{sec:bn}

As stated in \Cref{sec:opt}, we believe batch normalization layers are critical for learning discriminative features. As AdvProp \cite{advpropXie2020-md} assumes different underlying data distributions between actual data and generated examples, so that separating normal and adversarial data distributions could effectively improve the recognition performances, we hereby deploy auxiliary batch norms for the generated pseudo-anomaly data. Similarly, for the actual anomaly examples, we further trail on the effects of isolating anomaly data distributions from actual normalities, by switching off batch norm layers for true anomaly data. As shown in \Cref{tab:abl_bn}, the awareness of anomaly data distribution is critical under low-shot anomaly scenarios ($\gamma=.01$,$\gamma=.03$). Meanwhile, the affect of AdvProp and Freeze BN turns to be smaller if increasing the amount of anomaly data.

% \begin{table}[h]
%     \scriptsize
%     \centering
%     % \begin{adjustbox}{width=\linewidth,center}
%     \begin{tabular}{c l c c c c c}
%     \hline
%     Dataset &  Method & \multicolumn{4}{c}{Anomaly percentage $\gamma$}  \\
%         &   & 1 & 3 & 5 & 10 \\
%     \hline
%     \multirow{2}*{\makecell{ F-MNIST }}
%         & w/o & & {\bfseries 100.}±{\bfseries 0.00} & {\bfseries 100.}±{\bfseries 0.00}  \\
%         & w/  & {\bfseries 98.3}$\pm${\bfseries 2.21} & {\bfseries 99.9}±{\bfseries 0.03} & {\bfseries 100.}±{\bfseries 0.00} & {\bfseries 100.}±{\bfseries 0.00}  \\
%         \cline{2-6}
%     \multirow{2}*{\makecell{ CIFAR10 }}
%         & w/o & & & 99.4,1.14 \\
%         & w/  & {\bfseries 92.6}$\pm${\bfseries 7.79} & {\bfseries 96.2}±{\bfseries 5.19} & {\bfseries 97.2}±{\bfseries 3.35} & {98.6}±{2.58} \\
%         \cline{2-6}
%     \multirow{2}*{\makecell{ CIFAR100 }}
%         & w/o & & & & {\bfseries 99.7}$\pm${\bfseries 0.77} \\
%         & w/  & {\bfseries 97.3}$\pm${\bfseries 5.94} & & & {\bfseries 99.6}$\pm${\bfseries 1.11} \\
%     \hline
%     \end{tabular}
%     % \end{adjustbox}
%     \caption{Ablation study on batch normalization layers.}
%     \label{tab:abl_bn}
% \end{table}

\begin{table}[pos=h]
    \scriptsize
    \centering
    % \begin{adjustbox}{width=\linewidth,center}
    \setlength{\tabcolsep}{1pt}
    \begin{tabular}{c c c c c c c c c}
    \toprule
    Dataset &  Adv & Freeze & \multicolumn{5}{c}{Anomaly percentage $\gamma$}  \\
            &   Prop   &  BN   & .00 & .01 & .03 & .05 & .10 \\
    \midrule
    \parbox[t]{2mm}{\multirow{4}{*}{\rotatebox[origin=c]{90}{F-MNIST}}}
    % \multirow{4}*{\makecell{ F-MNIST }}
        & \checkmark & \checkmark & -  & {\bfseries 100.}±{\bfseries 0.00} & {\bfseries 100.}±{\bfseries 0.00} & {\bfseries 100.}±{\bfseries 0.00} & {\bfseries 100.}±{\bfseries 0.00}\\
        & \checkmark &            & {98.2}±{2.45} & {\bfseries 100.}±{\bfseries 0.00} & {\bfseries 100.}±{\bfseries 0.00} & {\bfseries 100.}±{\bfseries 0.00} & {\bfseries 100.}±{\bfseries 0.00} \\
        &            & \checkmark  & - & {98.2}$\pm${2.34} & {99.9}±{0.03} & {\bfseries 100.}±{\bfseries 0.00} & {\bfseries 100.}±{\bfseries 0.00} \\
        &            &   & {\bfseries 98.3}±{\bfseries 2.21} & {\bfseries 100.}±{\bfseries 0.00} & {\bfseries 100.}±{\bfseries 0.00} & {\bfseries 100.}±{\bfseries 0.00} & {\bfseries 100.}±{\bfseries 0.00}  \\
        \cmidrule{2-8}
    \parbox[t]{2mm}{\multirow{4}{*}{\rotatebox[origin=c]{90}{CIFAR10}}}
    % \multirow{4}*{\makecell{ CIFAR10 }}
        & \checkmark & \checkmark & - & {93.8}±{\bfseries 5.50} & {95.4}±{5.87}  & {99.2}±{1.55}  & {99.8}±{0.22} \\
        & \checkmark &   & {89.6}±{10.0} & {93.2}±{7.20} & {97.6}±{3.40}  & {\bfseries 99.4}±{\bfseries 1.08} & {99.7}±{0.69}  \\
        &            & \checkmark & - & {94.3}±{7.90} & {\bfseries 98.3}±{2.58} & {98.6}±{2.58} & {\bfseries 99.9}±{\bfseries 0.05} \\
        &            &  & {\bfseries 92.6}$\pm${\bfseries 7.76} & {\bfseries 94.8}±{6.17} & {\bfseries 98.3}±{\bfseries 2.38} & {99.3}±{1.15} & {\bfseries 99.9}±{0.15} \\
        \cmidrule{2-8}
    \parbox[t]{2mm}{\multirow{4}{*}{\rotatebox[origin=c]{90}{CIFAR100}}}
    % \multirow{4}*{\makecell{ CIFAR100 }}
        & \checkmark & \checkmark     & -                       & {96.0}$\pm${7.38} & {92.6}$\pm${11.3} & {95.6}±{9.24} & {98.8}±{3.27} \\
        & \checkmark &                & {92.4}±{\bfseries 11.4} & {95.7}$\pm${7.72} & {97.5}$\pm${5.59} & {98.1}±{5.47} & {99.1}$\pm${3.08} \\
        &            & \checkmark     & - & {96.0}$\pm${7.56} & { 97.3}$\pm${ 5.94} & {\bfseries 98.8}$\pm${\bfseries 2.82} & {99.6}$\pm${1.09} \\
        &            &                & {\bfseries 92.8}±{11.5} & {\bfseries 97.1}$\pm${\bfseries 6.20} & {\bfseries 97.8}$\pm${\bfseries 4.90} & {\bfseries 98.8}$\pm${3.63} & {\bfseries 99.7}$\pm${\bfseries 0.80} \\
    \bottomrule
    \end{tabular}
    % \end{adjustbox}
    \caption{Ablation study on different batch normalization strategy. We experimented on Fashion-MNIST, CIFAR10, and CIFAR100 datasets with different levels of anomaly involvement. Bold numbers represent the best results.}
    \label{tab:abl_bn}
\end{table}

% \begin{itemize}
%     \item anomaly loss (negative or reciprocal)
%     \item anomaly envolving (only collect partial anomaly classes)
% \end{itemize}

\subsection{Effectiveness of Anomalous Data}

Since our model achieves good AUCs if trained with more than 5\% anomaly data, we further investigat the benefits of increasing the anomaly data. Here, we evaluate the class-wise prediction anomaly score for two relatively complex datasets, CIFAR10 and CIFAR100, to examine the discrimination abilities between the models trained with 5\% and 20\% anomaly data respectively with box plots. To mitigate the side-effects of extreme values, \textit{Tukey's method} is adopted to remove the potential outlier scores outside of the interval $[Q_1-1.5IQR,Q_3+1.5IQR]$, where $Q_1$ and $Q_3$ are the first and third quartiles of the distribution and $IQR$ is the interquartiale range. As \Cref{fig:boxplot} shows, with more anomaly data, the model tends to have better discrimination on normal and anomaly data, as expected. Thus, though the marginal benefit might be low, we believe the increasing of anomaly data could contribute and potentially improve the model robustness.

\section{Discussion}

This work presented AGAD, an adversarial generative anomaly detection framework that fits for both supervised and semi-supervised anomaly detection tasks. In general, we proposed a anomaly detection paradigm taking the advantages of \textit{contextual adversarial information}, by learning discriminative features between normal and (pseudo-)abnormal data in a constrastive fashion. With extensive experiments, we found our method is performant with semi-supervised training protocol, while it gets more robust with a growing level of supervision. In such sense, we believe this proposed work fits towards most practical applications where anomaly data is collectable.

In general, our proposed method aims at addressing anomaly detection problems without or with limited anomaly data. Apart from the scarcity, the diversity of anomaly data is also worth being researched. Particularly, with only few types of anomalous data, how would it affect the anomaly detection performances? Also, other than assuming there is no anomaly data available, we believe the more practical problem is to know how many data need to be collected for real industrial applications. One future work direction can be to measure the sufficiency of anomalous data. Moreover, our initiative regarding developing a reconstruction-based method is due to its nature of visual explainability for anomalous data. Though qualitative analysis demonstrated different reconstruction directions towards normal and abnormal data, the reconstructed images are less expressive for detailed anomaly features. Future works may target at improving the visual explainability to disclose more fine-grained and critical anomaly features in a contrastive manner by taking the advantages \textit{contextual adversarial information}. Meanwhile, as a general paradigm, we also believe that our method can extend to other anomaly tasks (e.g. audio, ECG data).

% Further researches could explore the usability with more advanced generator (e.g. UNet++ \cite{Zhou2020}, Swin-UNet \cite{cao2021swinunet}), and loss functions (e.g. SSIM-L1 \cite{Zhao2017}), etc.

\bibliographystyle{cas-model2-names}
\bibliography{refs}

\clearpage

\appendix

\section{Additional Information}

In this section, the detailed network architectures and the index-class mappings of each benchmark dataset are provided.

\subsection{Index-class mapping}\label{sec:label_tag}

The corresponding index-class mapping for each benchmark dataset used in \Cref{sec:benchmark} is summarised in \Cref{tab:label_tag}.

\begin{table*}[pos=b]
    \begin{tabular}{c c | c c | c c c c}
    \hline
    \multicolumn{2}{c}{Fashion-MNIST} &  \multicolumn{2}{|c|}{CIFAR10} & \multicolumn{4}{c}{CIFAR100} \\
    \hline
    0 & T-shirt\&top & 0 & airplane    & 0 & aquatic mammals & 10  & large natural outdoor scenes\\
    1 & Trouser      & 1 & automobile  & 1 & fish & 11 & large omnivores and herbivores \\ 
    2 & Pullover     & 2 & bird        & 2 & flowers & 12 & medium-sized mammals \\
    3 & Dress        & 3 & cat         & 3 & food containers & 13 & non-insect invertebrates  \\
    4 & Coat         & 4 & deer        & 4 & fruit and vegetables & 14 & people \\
    5 & Sandal       & 5 & dog         & 5 & household electrical devices & 15 & reptiles \\
    6 & Shirt        & 6 & frog        & 6 & household furniture & 16 & small mammals \\
    7 & Sneaker      & 7 & horse       & 7 & insects & 17 & trees \\
    8 & Bag          & 8 & ship        & 8 & large carnivores & 18 & vehicles 1 \\
    9 & Ankle boot   & 9 & truck       & 9 & large man-made outdoor things & 19 & vehicles 2 \\
    \hline
    \end{tabular}
    \caption{Label index-class mapping.}
    \label{tab:label_tag}
\end{table*}

\subsection{Network architectures}
\label{sec:app_arch}

The network architectures used in this work include the naive encoder-decoder, UNet, and UNet++. The architecture is summarised as in \Cref{tab:naive_arch}. Meanwhile, we demonstrate the model size and running speed in \Cref{tab:gen}. The inference speed measured is the duration of inferencing one 32x32 image. Here, the naive encoder-decoder contains the smallest parameter sizes and has the fastest inference speed, while the UNet++ is 3 times slower than UNet though similar parameter sizes obtained.

\begin{table}[pos=h]
    \scriptsize
    \centering
    \begin{minipage}[t]{0.8\columnwidth}
    \begin{adjustbox}{width=\textwidth}
    \begin{tabular}{p{1.cm} | c c c }
    \toprule
    & Operation & ~~~in~~~ & ~~~out~~~ \\
    \midrule
    % kernel_size=(4,4), stride=(2, 2), padding=(1, 1)
    \multirow{9}*{Encoder}
        & Conv & 1 & 64 \\
        & LeakyReLU & - & - \\
        & Conv & 64 & 128 \\
        & BatchNorm & 128 & - \\
        & LeakyReLU & - & - \\
        & Conv & 128 & 256 \\
        & BatchNorm & 256 & - \\
        & LeakyReLU & - & - \\
        & Conv & 256 & 100 \\
    % \hline
    % \end{tabular}
    % \end{adjustbox}
    % \end{minipage}
    % \begin{minipage}[t]{0.8\columnwidth}
    % \begin{adjustbox}{width=\textwidth}
    % \begin{tabular}{p{1.cm} | c c c }
    % \hline
    % & Operation & ~~~in~~~ & ~~~out~~~ \\ 
    \midrule
    % kernel_size=(4,4), stride=(2, 2), padding=(1, 1)
    \multirow{9}*{Decoder}
        & ConvTranspose & 100 & 256 \\
        & BatchNorm & 256 & - \\
        & ReLU & - & - \\
        & ConvTranspose & 256 & 128 \\
        & BatchNorm & 128 & - \\
        & ReLU & - & - \\
        & ConvTranspose & 128 & 64 \\
        & BatchNorm & 64 & - \\
        & ReLU & - & - \\
        & ConvTranspose & 64 & 1 \\
    \bottomrule
    \end{tabular}
    \end{adjustbox}
    \end{minipage}
    \caption{The naive encoder-decoder network architecture. For each $Conv$ and $ConvTranspose$ operation, kernel size is set to 4 with 2 stride and 1 padding. LeakyReLU uses 0.2 for negative slope.}
    \label{tab:naive_arch}
\end{table}

\begin{table}[pos=h]
    \centering
    \setlength{\tabcolsep}{1pt}
    \begin{tabular}{l | l c  r}
    \toprule
    Generator~~   & Spec & No. Param & ~Inference speed \\
    \midrule
    Naive     & naive encoder-decoder & 2.04M &  2.67ms \\
    UNet      & skip-connection & 7.76M &  5.41ms \\
    UNet++~~  & dense skip pathways & 9.04M & 16.73ms \\
    \bottomrule
    \end{tabular}
    \caption{An overview of the experimented networks.}
    \label{tab:gen}
\end{table}

\begin{figure}[pos=h!]
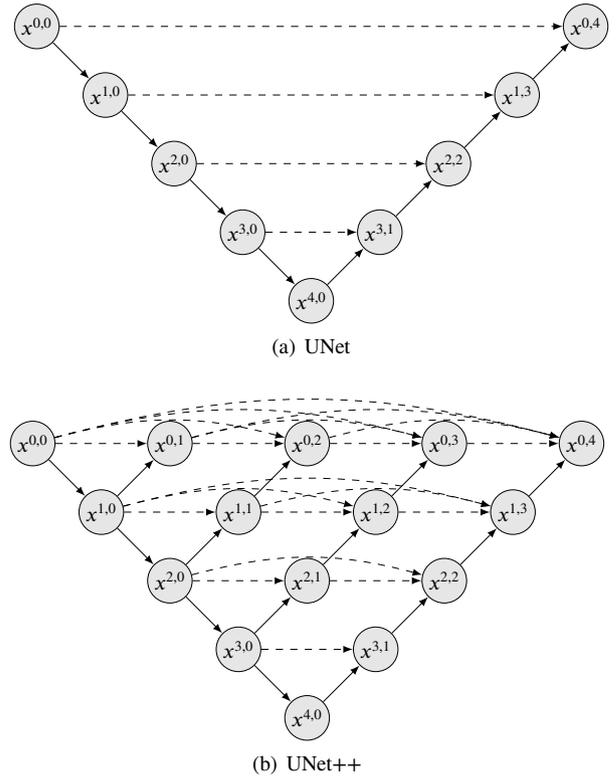

    \centering
    \subfigure[UNet]{
        \begin{minipage}[b]{0.45\textwidth}
        \begin{adjustbox}{width=\textwidth}
        \begin{tikzpicture}
            \import{.}{arch-unet.tex}
        \end{tikzpicture}
        \end{adjustbox}
        \end{minipage}
    }
    \qquad
    \subfigure[UNet++]{
        \begin{minipage}[b]{0.45\textwidth}
        \begin{adjustbox}{width=\textwidth}
        \begin{tikzpicture}
            \import{.}{arch-unetpp.tex}
        \end{tikzpicture}
        \end{adjustbox}
        \end{minipage}
    }
    \caption{Network Architectures. each node represents convolution operations (e.g. $ReLU-Conv-BatchNorm$ for downsampling, $ReLU-ConvTranspose-BatchNorm$ for upsampling), solid arrows mean forward operation, while dashed arrows mean skip-connections. Detailed network settings are as same as the original works.}
    \label{fig:app_arch}
\end{figure}

\begin{figure*}
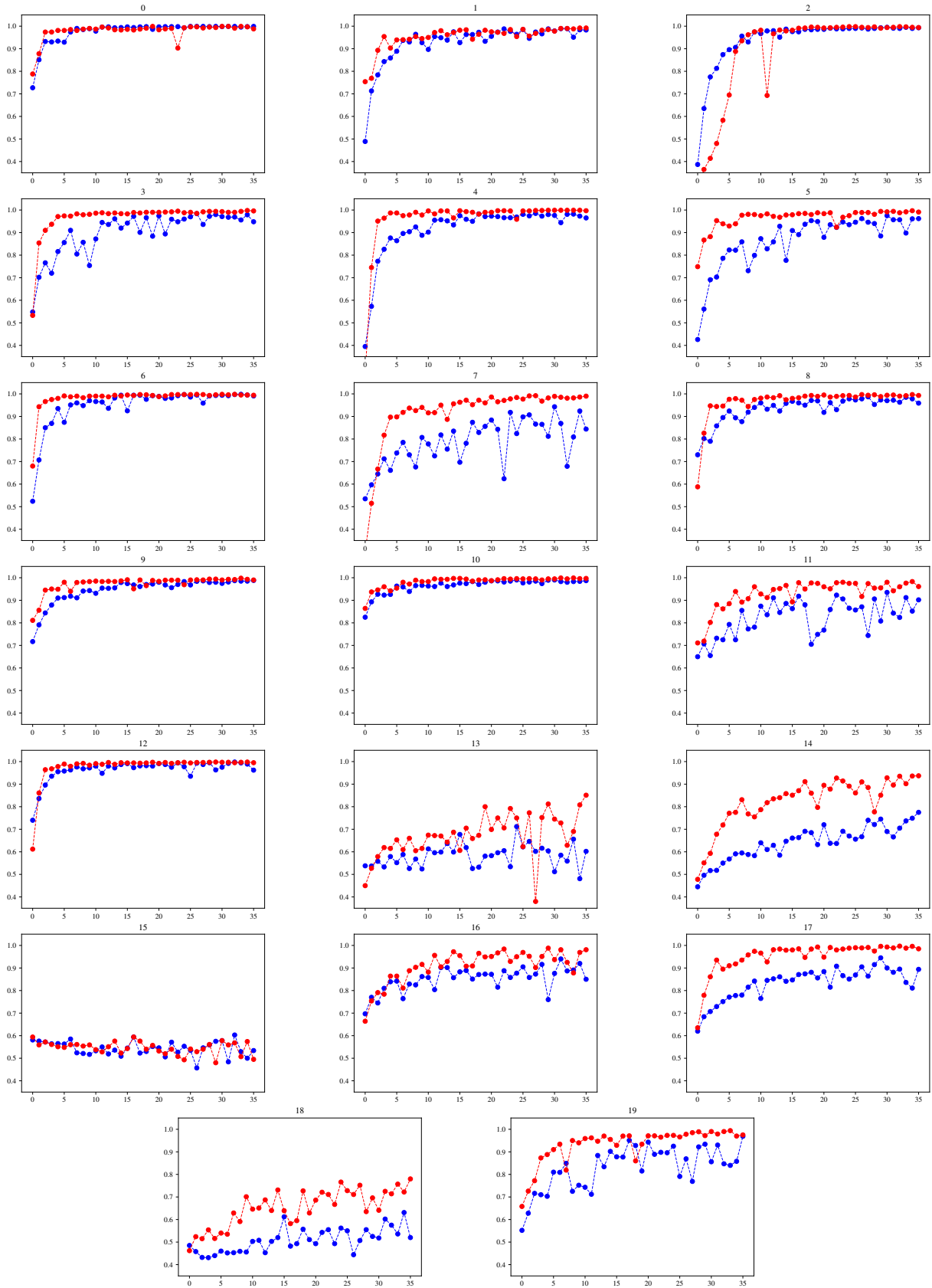

    \centering
    \begin{tabular}{c}
    \subfigure{
        \subfigure{
        \begin{minipage}[b]{.31\textwidth}
        \begin{adjustbox}{width=\linewidth,center}
            \input{plt-aquatic_mammals.pgf}
        \end{adjustbox}
        \end{minipage}
        }
        \subfigure{
        \begin{minipage}[b]{.31\textwidth}
        \begin{adjustbox}{width=\linewidth,center}
            \input{plt-fish.pgf}
        \end{adjustbox}
        \end{minipage}
        }
        \subfigure{
        \begin{minipage}[b]{.31\textwidth}
        \begin{adjustbox}{width=\linewidth,center}
            \input{plt-flowers.pgf}
        \end{adjustbox}
        \end{minipage}
        }
    }
    \\[-3.5em]
    \subfigure{
        \subfigure{
        \begin{minipage}[b]{.31\textwidth}
        \begin{adjustbox}{width=\linewidth,center}
            \input{plt-food_containers.pgf}
        \end{adjustbox}
        \end{minipage}
        }
        \subfigure{
        \begin{minipage}[b]{.31\textwidth}
        \begin{adjustbox}{width=\linewidth,center}
            \input{plt-fruit_and_vegetables.pgf}
        \end{adjustbox}
        \end{minipage}
        }
        \subfigure{
        \begin{minipage}[b]{.31\textwidth}
        \begin{adjustbox}{width=\linewidth,center}
            \input{plt-household_electrical_devices.pgf}
        \end{adjustbox}
        \end{minipage}
        }
    }
    \\[-3.5em]
    \subfigure{
        \subfigure{
        \begin{minipage}[b]{.31\textwidth}
        \begin{adjustbox}{width=\linewidth,center}
            \input{plt-household_furniture.pgf}
        \end{adjustbox}
        \end{minipage}
        }
        \subfigure{
        \begin{minipage}[b]{.31\textwidth}
        \begin{adjustbox}{width=\linewidth,center}
            \input{plt-insects.pgf}
        \end{adjustbox}
        \end{minipage}
        }
        \subfigure{
        \begin{minipage}[b]{.31\textwidth}
        \begin{adjustbox}{width=\linewidth,center}
            \input{plt-large_carnivores.pgf}
        \end{adjustbox}
        \end{minipage}
        }
    }
    \\[-3.5em]
    \subfigure{
        \subfigure{
        \begin{minipage}[b]{.31\textwidth}
        \begin{adjustbox}{width=\linewidth,center}
            \input{plt-large_man-made_outdoor_things.pgf}
        \end{adjustbox}
        \end{minipage}
        }
        \subfigure{
        \begin{minipage}[b]{.31\textwidth}
        \begin{adjustbox}{width=\linewidth,center}
            \input{plt-large_natural_outdoor_scenes.pgf}
        \end{adjustbox}
        \end{minipage}
        }
        \subfigure{
        \begin{minipage}[b]{.31\textwidth}
        \begin{adjustbox}{width=\linewidth,center}
            \input{plt-large_omnivores_and_herbivores.pgf}
        \end{adjustbox}
        \end{minipage}
        }
    }
    \\[-3.5em]
    \subfigure{
        \subfigure{
        \begin{minipage}[b]{.31\textwidth}
        \begin{adjustbox}{width=\linewidth,center}
            \input{plt-medium-sized_mammals.pgf}
        \end{adjustbox}
        \end{minipage}
        }
        \subfigure{
        \begin{minipage}[b]{.31\textwidth}
        \begin{adjustbox}{width=\linewidth,center}
            \input{plt-non-insect_invertebrates.pgf}
        \end{adjustbox}
        \end{minipage}
        }
        \subfigure{
        \begin{minipage}[b]{.31\textwidth}
        \begin{adjustbox}{width=\linewidth,center}
            \input{plt-people.pgf}
        \end{adjustbox}
        \end{minipage}
        }
    }
    \\[-3.5em]
    \subfigure{
        \subfigure{
        \begin{minipage}[b]{.31\textwidth}
        \begin{adjustbox}{width=\linewidth,center}
            \input{plt-reptiles.pgf}
        \end{adjustbox}
        \end{minipage}
        }
        \subfigure{
        \begin{minipage}[b]{.31\textwidth}
        \begin{adjustbox}{width=\linewidth,center}
            \input{plt-small_mammals.pgf}
        \end{adjustbox}
        \end{minipage}
        }
        \subfigure{
        \begin{minipage}[b]{.31\textwidth}
        \begin{adjustbox}{width=\linewidth,center}
            \input{plt-trees.pgf}
        \end{adjustbox}
        \end{minipage}
        }
    }
    \\[-3.5em]
    \subfigure{
        \subfigure{
        \begin{minipage}[b]{.31\textwidth}
        \begin{adjustbox}{width=\linewidth,center}
            \input{plt-vehicles_1.pgf}
        \end{adjustbox}
        \end{minipage}
        }
        \subfigure{
        \begin{minipage}[b]{.31\textwidth}
        \begin{adjustbox}{width=\linewidth,center}
            \input{plt-vehicles_2.pgf}
        \end{adjustbox}
        \end{minipage}
        }
        \subfigure{
        \begin{minipage}[b]{.31\textwidth}
        \end{minipage}
        }
    }
    \\[-1.5em]
    \end{tabular}
    \caption{Training curve of each class in CIFAR100. Red color represents UNet++, while blue color represents UNet, and x and y-axis denotes epoch and AUC respectively.}
    \label{fig:cifar100_curve}
\end{figure*}

\section{Additional Experiments}

This section briefly discussed some considerations on different backbones of generators.%, and presented a further evaluation regarding several real-world datasets.

\subsection{Choices of generator backbones }

We compared different backbone architectures of generators across the benchmark datasets. An overview of the chosen architectures is presented in \Cref{sec:app_arch}, and the experiment results are shown in \Cref{tab:oc_ssup_backbone}. Surprisingly, the naive generator significantly outperformed advanced frameworks like UNet or UNet++ for MNIST. We hereby replaced our UNet++ backbone generator with the naive encoder-decoder for MNIST dataset in the \Cref{sec:benchmark}.

% FYI: https://pubmed.ncbi.nlm.nih.gov/31841402/
% redesigning skip connections to aggregate features of varying semantic scales at the decoder sub-networks, leading to a highly flexible feature fusion scheme

As shown in \Cref{tab:oc_ssup_backbone}, the UNet \citeAP{UNet10.1007/978-3-319-24574-4_28} structure performed worst for Fashion-MNIST, and even failed for MNIST dataset with 31.7\% performance drop against the naive generator. As stated by \citeAP{Zhou2020}, the straight-away skip-connection of UNet  causes \textit{semantic gaps} between the feature maps of the encoder and decoder sub-networks. We suspect that the \textit{semantic gaps} would highly ease the feature representation learning when reconstructing simple data like MNIST. Since both normal and abnormal of simple data can be reconstructed easily, the model fails to detect anomalies based on reconstruction errors. UNet++ \citeAP{Zhou2020} addressed this problem by introducing \textit{dense skip pathways} for better feature learning , thus retaining discriminative reconstruction capability for normal and abnormal data. Presumably, by alleviating the \textit{semantic gaps}, UNet++ demonstrated smoother training curves compared to UNet as shown in \Cref{fig:cifar100_curve}.

\begin{table*}[pos=b]
    \begin{adjustbox}{width=.9\hsize,center}
    \begin{tabular}{c r c c c c c c c c c c c c }
    \toprule
    Dataset &  Generator & 0 & 1 & 2 & 3 & 4 & 5 & 6 & 7 & 8 & 9 & avg & SD   \\[-1ex]
    ------ & ------ & --- & --- & --- & --- & --- & --- & --- & --- & --- & --- & --- & --- \\[-1ex]
    \multirow{3}*{MNIST}
        & Naive & {\bfseries 100.} & {\bfseries 100.} & {\bfseries 99.0} & {\bfseries 98.6} & {\bfseries 99.5} & {\bfseries 97.2} & {\bfseries 99.6} & {\bfseries 99.8} & {\bfseries 95.8} & {\bfseries 99.2} & {\bfseries 99.1} & {\bfseries 0.86} \\
        & UNet & 76.0 & 58.6 & 81.3 & 93.3 & 41.9 & 46.8 & 98.9 & 45.2 &  89.1 &  42.8 &  67.4 &  21.6 \\
        & UNet++ & 95.5 & 100. & 85.4 & 78.5 & 91.4 & 92.5 & 73.6 &  90.2 &  89.6 &  89.4 &  89.4 &  7.46 \\
        \cmidrule{2-14}
    \multirow{4}*{\makecell{ Fashion- \\ MNIST }}
        &  Generator & 0 & 1 & 2 & 3 & 4 & 5 & 6 & 7 & 8 & 9 & avg & SD   \\[-1ex]
        & ------ & --- & --- & --- & --- & --- & --- & --- & --- & --- & --- & --- & --- \\[-1ex]
        & Naive & 98.2 & 100. & 99.1 & 99.7 & 99.0 & 100. & 97.7 &  100. &  99.1 &  100. &  99.2 &  0.75 \\
        & UNet & {95.5} & {99.8} & {99.6} & {97.1} & {99.9} & {99.9} & {93.2} & {\bfseries 100.} & {99.4} & {99.0} & {98.3} & {2.21}  \\
        & UNet++ & {\bfseries 99.9} & {\bfseries 100.} & {\bfseries 100.} &  {\bfseries 99.4} & {\bfseries 100.} & {\bfseries 100.} & {\bfseries 100.} & {\bfseries 100.} & {\bfseries 100.} & {\bfseries 100.} & {\bfseries 99.9} & {\bfseries 0.18} \\
        \cmidrule{2-14}
    \multirow{4}*{\makecell{ ~~~CIFAR10~~~ }}
        &  Generator & 0 & 1 & 2 & 3 & 4 & 5 & 6 & 7 & 8 & 9 & avg & SD  \\[-1ex]
        & ------ & --- & --- & --- & --- & --- & --- & --- & --- & --- & --- & --- & --- \\[-1ex]
        & Naive & 79.9 & 59.1 & 64.9 & 60.1 &  68.9 &  69.8 &  88.2 &  61.3 &  83.0 &  74.3 &  71.0 &  9.63 \\
        & UNet & {\bfseries 99.9} & {88.3} & {89.8} & {94.1} & {99.3} & {87.7} & {99.8} & 73.8 & {\bfseries 99.9} & 92.3 & 92.5 & 7.81 \\
        & UNet++ & {\bfseries 99.9} & {\bfseries 99.5} & {\bfseries 98.8} & {\bfseries 98.4} & {\bfseries 99.7} & {\bfseries 94.7} & {\bfseries 100.} & {\bfseries 93.2} & {\bfseries 99.9} & {\bfseries 98.8} & {\bfseries 98.3} & {\bfseries 2.26} \\
    \midrule
    \end{tabular}
    \end{adjustbox}

    \begin{adjustbox}{width=.9\hsize,center}
    \begin{tabular}{c r c c c c c c c c c c c}
    \multirow{9}*{~~CIFAR100~~}
        &  Generator & 0 & 1 & 2 & 3 & 4 & 5 & 6 & 7 & 8 & 9 & 10  \\[-1ex]
        & ------ & --- & --- & --- & --- & --- & --- & --- & --- & --- & --- & --- \\[-1ex] 
        & Naive & 64.3 & 90.2 & 77.8 & 82.0 & 67.3 & 62.7 & 71.4 & 77.1 & 68.5 & 82.8 & 89.3  \\
        & UNet & {\bfseries 99.9} & {99.4} & {99.7} & {99.3} & {98.9} & {97.5} & {99.8} & {96.0} & {\bfseries 99.4} & {99.3} & {99.3} \\
        & UNet++ & {\bfseries 99.9} & {\bfseries 99.6} & {\bfseries 99.8} & {\bfseries 99.8} & {\bfseries 99.9} & {\bfseries 99.7} & {\bfseries 99.9} & {\bfseries 99.9} & {\bfseries 99.4} & {\bfseries 99.8} & {\bfseries 99.9} \\
        \cmidrule{2-13}
        &  Generator & 11 & 12 & 13 & 14 & 15 & 16 & 17 & 18 & 19 & avg & SD  \\[-1ex]
        & ------ & --- & --- & --- & --- & --- & --- & --- & --- & --- & --- & --- \\[-1ex]
        & Naive & 56.7 & 69.5 & 73.3 & 63.7 & 65.0 & 67.4 & 81.2 & 61.0 & 70.7 & 71.6 & {\bfseries 8.87} \\
        & UNet & {96.3} & {99.8} & {74.7} & 78.9 & {\bfseries 62.6} & {95.6} & {95.5} & 66.8 & {96.9} & {92.8} & 11.5 \\
        & UNet++ & {\bfseries 99.1} & {\bfseries 99.9} & {\bfseries 86.4} & {\bfseries 96.9} & 61.7 & {\bfseries 99.0} & {\bfseries 99.8} & {\bfseries 86.5} & {\bfseries 99.7} & {\bfseries 96.3} & 8.88 \\
    \bottomrule
    \end{tabular}
    \end{adjustbox}
    \caption{One-class semi-supervised anomaly detection benchmark performances. We reported the average AUC in \% that is computed over 3 runs. Bold numbers represent the best results.}
    \label{tab:oc_ssup_backbone}
\end{table*}

As a result, for the simple datasets without complex patterns, the naive generator might be a yet best solution for relatively good performances as well as its smaller parameter size and faster inferencing speed, While UNet++ may fit more complex scenarios that need more advanced feature extraction.

\bibliographystyleAP{cas-model2-names}
\bibliographyAP{refs}






% \end{thebibliography}

\end{document}